\documentclass[letterpaper]{article} 
\usepackage{aaai24}  
\usepackage{times}  
\usepackage{helvet}  
\usepackage{courier}  
\usepackage[hyphens]{url}  
\usepackage{graphicx} 
\usepackage{natbib}  
\usepackage{caption} 
\usepackage{algorithm}
\usepackage{algorithmic}

\usepackage{amsmath}
\usepackage{amssymb}
\usepackage{subfigure}
\usepackage{multirow}

\usepackage{newfloat}
\usepackage{listings}
\DeclareCaptionStyle{ruled}{labelfont=normalfont,labelsep=colon,strut=off} 
\floatstyle{ruled}
\newfloat{listing}{tb}{lst}{}
\floatname{listing}{Listing}
\title{CF-NeRF: Camera Parameter Free Neural Radiance Fields with Incremental Learning}
\author{
    Qingsong Yan \textsuperscript{\rm{1}}, 
    Qiang Wang \textsuperscript{\rm{2,\thanks{Corresponding author}}}, 
    Kaiyong Zhao \textsuperscript{\rm{3}}, 
    Jie Chen \textsuperscript{\rm{4}}, \\
    Bo Li \textsuperscript{\rm{5}}, 
    Xiaowen Chu \textsuperscript{\rm{6,5,\footnotemark[1]}}, 
    Fei Deng \textsuperscript{\rm{1,7}}
}

\affiliations{
    \textsuperscript{1} Wuhan University, Wuhan, China \\
	\textsuperscript{2} Harbin Institute of Technology (Shenzhen), Shenzhen, China\\
	\textsuperscript{3} XGRIDS, Shenzhen, China \\
    \textsuperscript{4} Hong Kong Baptist University, Hong Kong SAR, China\\
    \textsuperscript{5} The Hong Kong University of Science and Technology, Hong Kong SAR, China\\
	\textsuperscript{6} The Hong Kong University of Science and Technology (Guangzhou), Guangzhou, China\\
    \textsuperscript{7} Hubei Luojia Laboratory, Wuhan, China\\
	{\tt\small  yanqs\_whu@whu.edu.cn, qiang.wang@hit.edu.cn, kyzhao@xgrids.com, chenjie@comp.hkbu.edu.hk } \\
	{\tt\small bli@cse.ust.hk, xwchu@ust.hk, fdeng@sgg.whu.edu.cn }

}

\usepackage{bibentry}

\begin{document}

\maketitle

\begin{abstract}

Neural Radiance Fields (NeRF) have demonstrated impressive performance in novel view synthesis. 
However, NeRF and most of its variants still rely on traditional complex pipelines to provide extrinsic and intrinsic camera parameters, such as COLMAP.
Recent works, like NeRFmm, BARF, and L2G-NeRF, directly treat camera parameters as learnable and estimate them through differential volume rendering. 
However, these methods work for forward-looking scenes with slight motions and fail to tackle the rotation scenario in practice.
To overcome this limitation, we propose a novel \underline{c}amera parameter \underline{f}ree neural radiance field (CF-NeRF), which incrementally reconstructs 3D representations and recovers the camera parameters inspired by incremental structure from motion (SfM). 
Given a sequence of images, CF-NeRF estimates the camera parameters of images one by one and reconstructs the scene through initialization, implicit localization, and implicit optimization. 
To evaluate our method, we use a challenging real-world dataset NeRFBuster which provides 12 scenes under complex trajectories. 
Results demonstrate that CF-NeRF is robust to camera rotation and achieves state-of-the-art results without providing prior information and constraints.

\end{abstract}

\section{Introduction}

3D reconstruction is a hot topic in computer vision that aims to recover 3D geometry from RGB images. 
However, traditional methods contain lots of complex procedures, such as feature extraction and matching \cite{lowe2004distinctive,yi2016lift}, sparse reconstruction \cite{agarwal2011building,wu2013towards,schonberger2016structure,moulon2017openmvg}, and dense reconstruction \cite{yao2018mvsnet,mi2022generalized,yan2022rethinking}. 
Consequently, traditional methods are not a differential end-to-end reconstruction pipeline and require high-quality results from each sub-module to achieve accurate results. When the quality of results is poor, it is challenging to identify which module is causing the problem.

\begin{figure} [!t]
    \centering
    
    \subfigure[NeRFmm]{
    \begin{minipage}[b]{0.27\linewidth}
    \includegraphics[width=1\linewidth]{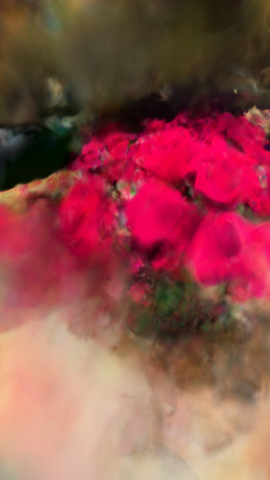}
    \end{minipage}}
    \subfigure[SiRENmm]{
    \begin{minipage}[b]{0.27\linewidth}
    \includegraphics[width=1\linewidth]{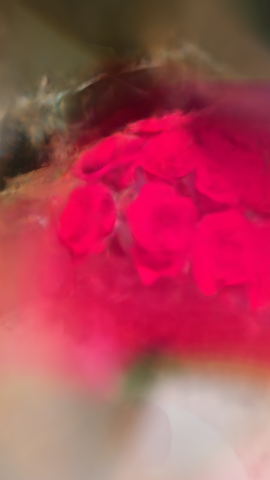}
    \end{minipage}}
    \subfigure[BARF]{
    \begin{minipage}[b]{0.27\linewidth}
    \includegraphics[width=1\linewidth]{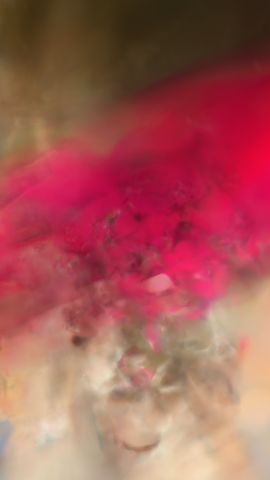}
    \end{minipage}}
    
    \vspace{-1.0 em}

    \subfigure[GARF]{
    \begin{minipage}[b]{0.27\linewidth}
    \includegraphics[width=1\linewidth]{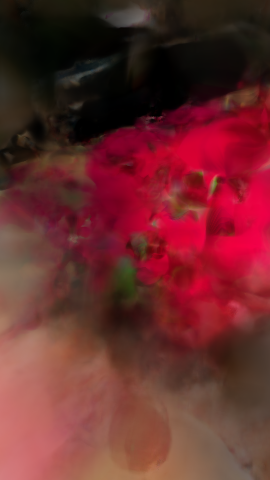}
    \end{minipage}}
    \subfigure[L2G-NeRF]{
    \begin{minipage}[b]{0.27\linewidth}
    \includegraphics[width=1\linewidth]{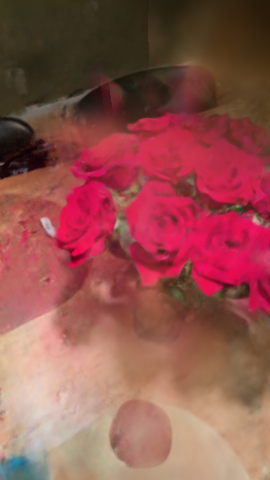}
    \end{minipage}}
    \subfigure[CF-NeRF]{
    \begin{minipage}[b]{0.27\linewidth}
    \includegraphics[width=1\linewidth]{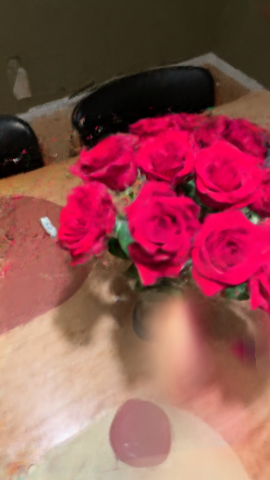}
    \end{minipage}}

    \vspace{-1.0 em}
    
    \caption{
		\textbf{Rendering results on the NeRFBuster} 
        We select an image sequence from NeRFBuster \cite{warburg2023nerfbusters} and use novel views synthesis to compare the quality of camera parameters from NeRFmm \cite{wang2021nerf}, SiRENmm \cite{ventusff2021}, BARF \cite{lin2021barf}, GARF \cite{chng2022garf}, L2G-NeRF \cite{chen2023local} and our method CF-NeRF.
	}

    \label{fig:com}
    \vspace{-1.5 em}
 
\end{figure}

Recently, Neural Radiance Fields (NeRF) \cite{mildenhall2021nerf,muller2022instant,yu2021plenoctrees,barron2021mip} has demonstrated a novel way to render highly realistic novel views with impressive quality. Without recovering 3D geometry, NeRF relies on multi-layer perception (MLP) to predict color and sigma for each point in the scene and samples several points along a ray to render a pixel through differential volume rendering. Unlike traditional 3D reconstruction, NeRF simplifies the reconstruction into one step and implicitly represents the 3D scene. Benefiting from the excellent ability of NeRF, it has been further extended to dynamic scenes \cite{pumarola2021d}, large-scale \cite{turki2022mega}, and even surface \cite{wang2021neus} and material reconstruction \cite{boss2021nerd}.

Despite the remarkable performance of NeRF and its variants in novel view synthesis,  they still require camera parameters before training. 
The most common processing pipeline is first recovering camera parameters using traditional complex methods like COLMAP \cite{schonberger2016structure} or OpenMVG \cite{moulon2017openmvg}, and then training the NeRF through differential volume rendering. 
In other words, the differentiability of the whole reconstruction pipeline is destroyed and divided into two separate parts, resulting in the NeRF not being end-to-end and the reconstruction quality being unidirectionally dependent on traditional methods.

To unify camera parameter estimation and reconstruction, researchers have tried to recover or optimize camera parameters along with NeRF \cite{wang2021nerf,lin2021barf,chng2022garf,kuang2022neroic,tancik2023nerfstudio}. The straightforward idea is to treat camera parameters as learnable. NeRFmm \cite{wang2021nerf} proves that it is possible to estimate camera parameters by NeRF. BARF \cite{lin2021barf} recovers extrinsic camera parameters and the NeRF model by dynamically adjusting weights of different frequencies of positional encoding. GARF \cite{chng2022garf} replaces ReLU with Gaussian activations to obtain high-accuracy results. NeROIC \cite{kuang2022neroic} and NeRFStudio \cite{tancik2023nerfstudio} optimize camera parameters obtained from COLMAP and the NeRF simultaneously. However, these methods are only suitable for forward-looking scenes or scenes with initial camera parameters and cannot be directly used in the real world with complex movement.

This paper proposes a new end-to-end approach called camera parameter free NeRF (CF-NeRF) to address the limitations of existing NeRF-based methods in estimating camera parameters.
Figure \ref{fig:com} compares rendered novel views by camera parameters estimated by NeRFmm \cite{wang2021nerf}, SiRENmm \cite{ventusff2021}, BARF \cite{lin2021barf}, GARF \cite{chng2022garf}, L2G-NeRF \cite{chen2023local}, and our method CF-NeRF, where CF-NeRF is the only method that successfully reconstructs the 3D scene with rotation. 
Unlike other methods that simultaneously estimate all camera parameters, CF-NeRF inherits ideas from incremental 
structure from motion (SfM) and recovers camera parameters one by one. CF-NeRF contains three major components: initialization, implicit localization, and implicit optimization. 
CF-NeRF first uses initialization to recover the camera parameters and NeRF by a few images and then gradually estimates the camera parameters of other images through two steps:  the implicit localization provides an initial camera parameter for the newly added image, and the implicit optimization optimizes the camera parameter of all images to reduce drift. Our contributions are as follows:
\begin{enumerate}
	\item We propose a novel end-to-end method, CF-NeRF, that does not need prior information or constraints to recover the intrinsic and extrinsic camera parameters and the NeRF simultaneously.
	\item We design an incremental training pipeline for the CF-NeRF, inspired by the incremental SfM, to avoid trapping to local minimal and is suitable for complex trajectories.
	\item Experiments of our method achieve state-of-the-art results on the NeRFBuster dataset \cite{warburg2023nerfbusters} captured in the real world, proving that the CF-NeRF can estimate accurate camera parameters with the specifically designed training procedure.
\end{enumerate}

\section{Related Work}

In this section, we first introduce the development of NeRF-related methods with known camera parameters. 
Then, we describe several camera parameter estimation methods using traditional SfM\&SLAM (simultaneous localization and mapping) and the NeRF.

\subsection{NeRF}

NeRF \cite{mildenhall2021nerf} uses the MLP to represent the 3D scene implicitly and can be trained through differential volume rendering from a set of images with known camera parameters.
However, NeRF suffers from efficiency and needs around 1-2 days to train a scene and several minutes to render a novel view at the testing. Instant-NGP \cite{muller2022instant} builds a multi-resolution hash table to store space-aware feature vectors and reduces the complexity of the MLP network. 
Meanwhile, \cite{sun2022direct,fridovich2022plenoxels,zhang2022digging} try to use the coarse-to-fine strategy and \cite{yu2021plenoctrees,chen2022tensorf,garbin2021fastnerf,wadhwani2022squeezenerf} update the network structure to speed up training or testing.
Besides, NeRF faces another problem that it cannot work for large-scale, unbounded 3D scenes. NeRF++ \cite{zhang2020nerf++} and MipNeRF360 \cite{barron2021mip, barron2022mip} utilize different sampling strategies for foreground and background to model unbounded 3D scenes by a finite volume. MegaNeRF \cite{turki2022mega} and BlockNeRF \cite{tancik2022block} split a large scene into multiple small regions and assign a network for each part.
Moreover, \cite{martin2021nerf,pumarola2021d,park2021hypernerf,attal2021torf} extend NeRF to dynamic scenes and \cite{jain2021putting,yu2021pixelnerf,niemeyer2022regnerf,kim2022infonerf} introduce context or geometry information into NeRF to suit scenes with sparse views.
In addition to the advances in novel view synthesis, NeRF has made significant progress in geometric reconstruction \cite{yariv2021volume,wang2022hf,darmon2022improving,long2022neuraludf,long2022sparseneus,fu2022geo}. UniSURF \cite{oechsle2021unisurf} and NeUS \cite {wang2021neus} estimate the zero-level set of an implicit signed distance function instead of the space density. Furthermore, some work \cite{zhang2021nerfactor,verbin2022ref,boss2021nerd,kuang2022neroic,boss2021neural,boss2022samurai} even combines BRDF and NeRF to decompose a scene into shape, reflectance, and illumination.
However, all of these methods split the reconstruction into two steps and require traditional methods to provide camera parameters, which significantly limits the application of NeRF.

\begin{figure*}[t]

    \begin{center}
       \includegraphics[width=\linewidth]{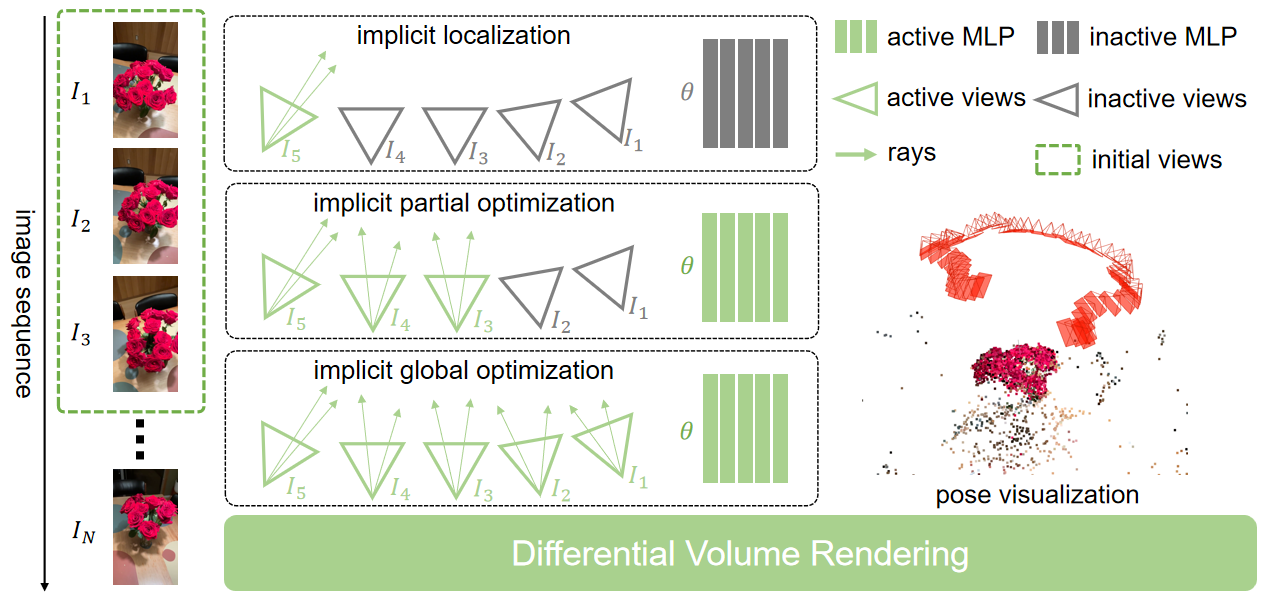}
    \end{center}

    \vspace{-0.5em}
    \caption{
        \textbf{The pipeline of CF-NeRF} CF-NeRF can estimate the weight $\theta$ of NeRF $\mathcal{F}$ and the camera parameter $\delta$. After initializing through a few selected images, CF-NeRF recovers $\delta$ of the image one by one through implicit localization that only optimizes the newly added image and implicit optimization that refines $\theta$ and $\delta$. Implicit optimization can be divided into partial and global optimization depending on the number of images used. We visualize $\delta$ reconstructed by CF-NeRF and sparse points from COLMAP \cite{schonberger2016structure} to show that CF-NeRF can reconstruct rotation in image sequences.
        }

    \label{fig:pipeline}
    \vspace{-1.0em}
\end{figure*}

\subsection{Camera Parameter Estimation}

Traditional SfM \cite{agarwal2011building,wu2013towards,chatterjee2013efficient,moulon2013global,schonberger2016structure,moulon2017openmvg} and SLAM \cite{mur2015orb,engel2014lsd,sumikura2019openvslam,engel2017direct} can estimate camera parameters for given images. 
However, these methods divide the reconstruction pipeline into several different modules that need hand-crafted features \cite{lowe2004distinctive,bay2006surf} or learning-based methods \cite{yi2016lift,detone2018superpoint} or the optical flow \cite{lucas1981iterative,teed2020raft} to establish image correspondences,  and
then reconstruct a sparse scene and camera parameters through several separate non-differentiable steps.

In light of these limitations, it is worth exploring to estimate camera parameters during the training process of NeRF.
The most direct attempt to utilize NeRF is the visual localization, where iNeRF \cite{yen2021inerf}, NeDDF \cite{ueda2022neural}, and PNeRFP \cite{lin2022parallel} try to estimate the extrinsic camera parameter of a new image by a pre-trained NeRF model. 
Then, NeRFmm \cite{wang2021nerf} and SiRENmm \cite{ventusff2021} take the NeRF and camera parameters as learnable and prove that it is possible to train the NeRF model from scratch without camera parameters, but they only work for forward-looking scenes. 
To further enhance accuracy in forward-looking or rotation scenes with initial camera parameters, BARF \cite{lin2021barf} dynamically adjusts the weight of the positional encoding, GARF \cite{chng2022garf} replaces the ReLU activate function with the Gaussian activation function, and L2G-NeRF \cite{chen2023local} introduces a local-to-global registration. 
Interestingly, GNeRF \cite{meng2021gnerf} and VMRF \cite{zhang2022vmrf} assume there is a prior known distribution of camera parameters to decrease the freedom of camera parameters during training the NeRF model.
Meanwhile, other researchers try to add different external restrictions to guide the camera parameter estimation. SCNeRF \cite{jeong2021self} and Level-$S^2$fM \cite{xiao2023level} rely on feature matches to guide camera parameters estimation. NoPe-NeRF \cite{bian2022nope}, iMap \cite{sucar2021imap}, NeRF-SLAM \cite{rosinol2022nerf}, Nice-SLAM \cite{zhu2022nice}, and Nicer-SLAM \cite{zhu2023nicer} integrate depth maps from active sensors or CNN networks to tune the NeRF. Additionally, LocalLR \cite{meuleman2023progressively} combines depth maps and optical flow to train NeRF.

Regrettably, images acquired from real-world scenarios often exhibit a multitude of challenges. These challenges include rotations and the absence of prior information of camera parameters. Furthermore, the introduction of external constraints can augment the intricacy and unpredictability of the reconstruction process.
To solve these problems, we propose CF-NeRF inspired by the traditional incremental SfM, which does not require any prior information or external constraints while reconstructing the 3D scene and camera parameters end-to-end from image sequences, demonstrating the powerful reconstruction capability of the NeRF after using a specific training strategy.

\section{Method}

In this section, we provide an overview of the proposed method. Firstly, we introduce the preliminary background of the NeRF and the traditional incremental SfM. Then, we explain the details of CF-NeRF that can recover camera parameters from image sequences.

\subsection{Preliminary Background}

\begin{figure}[t]
    \begin{center}
       \includegraphics[width=0.8\linewidth]{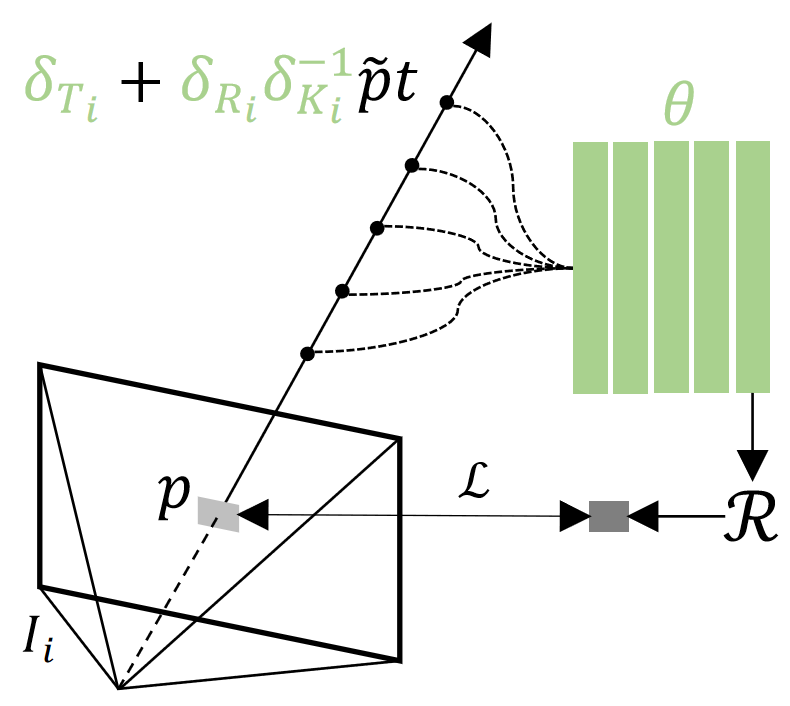}
    \end{center}

    \vspace{-1.0em}
    
    \caption{
        \textbf{Estimated Parameters} CF-NeRF estimates the weight $\theta$ of NeRF model and the camera parameter $\delta$, which include the camera rotation $\delta_R$, the camera translation $\delta_T$, and camera intrinsic parameter $\delta_K$.
        } 
    \label{fig:ire_parameter}

    \vspace{-1.5em}
    
\end{figure}

\paragraph{NeRF}

NeRF can generate realistic images from a set of images $ I= ( I_1, I_2, ..., I_N )$ from $N$ different places without explicitly reconstructing. However, NeRF needs associated camera parameters $\delta$, including camera rotation $\delta_R=(\delta_{R_1},\delta_{R_2},...,\delta_{R_N})$, camera translation $\delta_T=(\delta_{T_1},\delta_{T_2},...,\delta_{T_N})$, and intrinsic camera parameter $\delta_K$.
Given a NeRF model $\mathcal{F}$ and corresponding weight $\theta$  , it can estimate color $c$ and density $\sigma$ through a implicit function $c(x,\vec{d}),\sigma(x)=\mathcal{F}_\theta(x,\vec{d})$ with a point $x$ and a view direction $\vec{d}$. To render a pixel $p$, NeRF needs to sample several points  $x_p(t)=o+ \vec{d}t$ along a ray shooting from the view position $o$ and generate the color $c_p$ by the volume rendering function $\mathcal{R}$ as Eq. \ref{eq:rend} shows, where $\mathcal{T}(t)=exp(-\int_{t_n}^t\sigma(x_p(s))ds)$ indicates the accumulated transmittance along the ray. $t_n$ and $t_f$ are the near and far bounds of the ray.

\begin{align}
    \label{eq:rend}
	c_p = \mathcal{R}(p|\theta) = \int_{t_n}^{t_f}\mathcal{T}(t)\sigma(x_p(t))c(x_p(t),\Vec{d})dt
\end{align}

Benefiting from the differential property of the volume rendering, NeRF can be trained end to end by minimizing the difference between $c_p$ and observed color $I(p)$ as Eq. \ref{eq:nerf_op} shows, where $\mathcal{L}$ is the loss function. To be noted, NeRF only estimates $\theta$ and borrows $\delta$ from traditional SfM methods. However, NeRFmm \cite{wang2021nerf} prove that it is possible to estimate $\theta$ and $\delta$ simultaneously under the forward-looking situation.

\begin{align}
    \label{eq:nerf_op}
	{ \underset{\theta}{{\arg\min}\,}  \{ \sum_{I_i \in I} \sum_{p \in I_i}{  \mathcal{L} ( \mathcal{R}(p|\theta), I_i(p) ) } \} }
\end{align}

\paragraph{Incremental SfM}

Given a set of images, the incremental SfM can recover $\delta$ one by one in a linear time \cite{wu2013towards} and contains four steps \cite{schonberger2016structure}:

\textbf{Initialization} The selection of an initial two-view is essential because a suitable initial two-view improves the robustness and quality of the reconstruction. With a given two-view and its matched features, incremental SfM computes the relative pose by multi-view geometry (MVG) and triangulates 3D points to initial the scene.

\textbf{Image Registration} After initialization, incremental SfM adds images to the scene in order. Given a new image, incremental SfM builds the 2D-3D relationship by matching its features with images in the scene and recovers the camera parameter by Perspective-n-Point (PnP).

\textbf{Triangulation} As a newly added image observes additional information that can extend the scale of the scene, incremental SfM triangulates more 3D points based on the new image and matched features.

\textbf{Bundle Adjustment} Adding new images and 3D points without refinement leads to drift. Therefore, it is essential to apply bundle adjustment (BA) by minimizing the re-projection error. In terms of efficiency, incremental SfM proposes partial BA that refines only a subset of images, and global BA that optimizes all images.

\subsection{CF-NeRF}

Fusing the differentiability of NeRF and the reconstruction strategy of SfM, we propose CF-NeRF, which is capable of estimating the camera parameter under complex movement from sequential images. CF-NeRF consists of three modules: initialization, implicit localization, and implicit optimization, as Figure \ref{fig:pipeline} shows. To convenient later introduction, we define the set of images we have completed estimating the camera parameter as $E$, which starts from $\varnothing$.

\paragraph{Parameter}

CF-NeRF estimates the camera parameter $\delta$ and the weight $\theta$ of NeRF, as Figure \ref{fig:ire_parameter} shows. The $\delta$ includes $\delta_R$, $\delta_T$, and $\delta_K$. During the differential volume rendering, we calculate the ray $\vec{r}_p(t)=\delta_{T_i}+\delta_{R_i}\delta_{K_i}^{-1}\tilde{p}t$ of pixel $p$ in image $I_i \in I$, where $\tilde{p}$ is the homogeneous expression of $p$.  
Following NeRFmm \cite{wang2021nerf}, we use the axis-angle to represent $\delta_R$ and assume all images have the same camera intrinsic parameter without distortion so that $\delta_K$ only contains the focal length.
We initialize $\delta_R$ and $\delta_T$ to zero, and set $\delta_K$ to $53^{\circ}$ by a common field of view. 
The $\theta$ is the weight of $\mathcal{F}$ that implicitly represents the scene. The activation function determines how to initialize $\theta$. NeRF using ReLU are initialized according to NeRF \cite{mildenhall2021nerf}, while NeRF using sine are initialized according to SIREN \cite{sitzmann2020implicit}.

\begin{figure*}
    \centering
    
    \subfigure[GT]{
    \begin{minipage}[b]{0.13\linewidth}
    \includegraphics[width=1\linewidth]{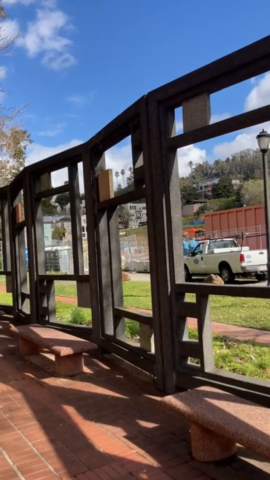} 
    \includegraphics[width=1\linewidth]{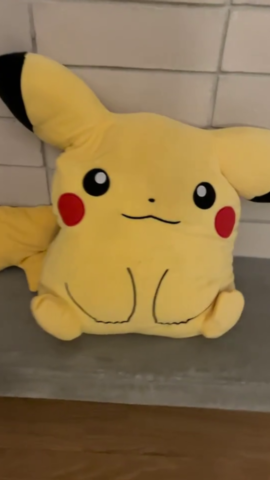}
    \end{minipage}}
    \subfigure[NeRFmm]{
    \begin{minipage}[b]{0.13\linewidth}
        \includegraphics[width=1\linewidth]{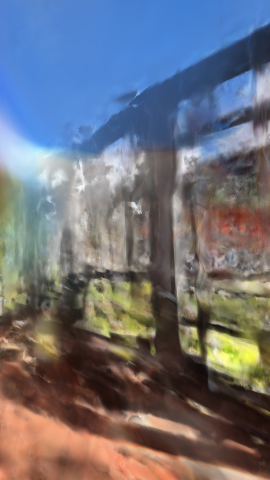}
        \includegraphics[width=1\linewidth]{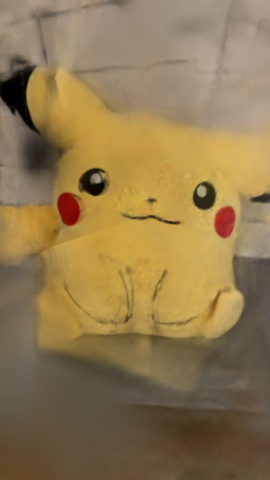} 
    \end{minipage}}
    \subfigure[SiRENmm]{
    \begin{minipage}[b]{0.13\linewidth}
        \includegraphics[width=1\linewidth]{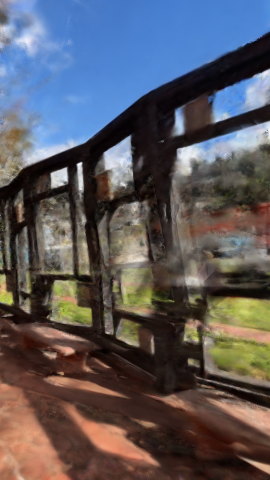}
        \includegraphics[width=1\linewidth]{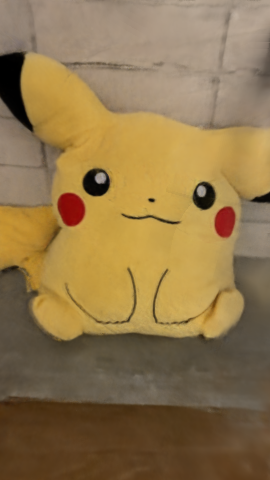}
    \end{minipage}}
    \subfigure[BARF]{
    \begin{minipage}[b]{0.13\linewidth}
        \includegraphics[width=1\linewidth]{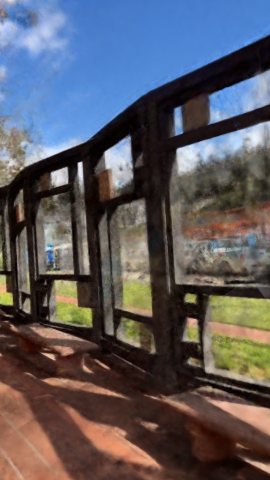}
        \includegraphics[width=1\linewidth]{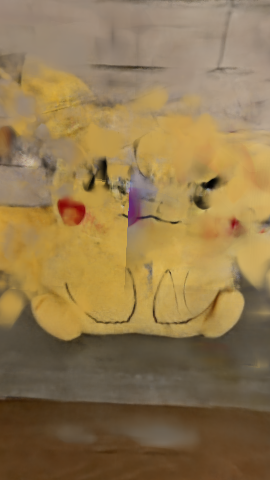}
    \end{minipage}}
    \subfigure[GARF]{
    \begin{minipage}[b]{0.13\linewidth}
        \includegraphics[width=1\linewidth]{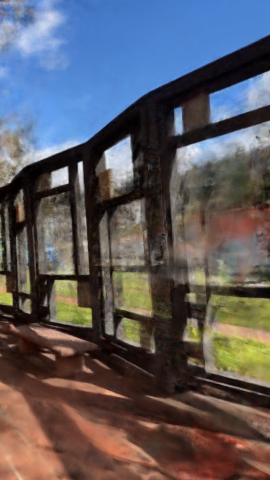}
        \includegraphics[width=1\linewidth]{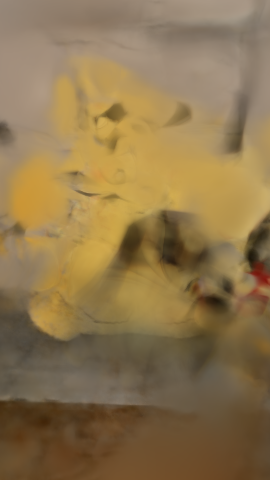}
    \end{minipage}}
    \subfigure[L2G-NeRF]{
    \begin{minipage}[b]{0.13\linewidth}
        \includegraphics[width=1\linewidth]{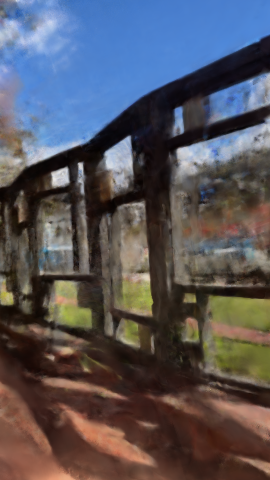}
        \includegraphics[width=1\linewidth]{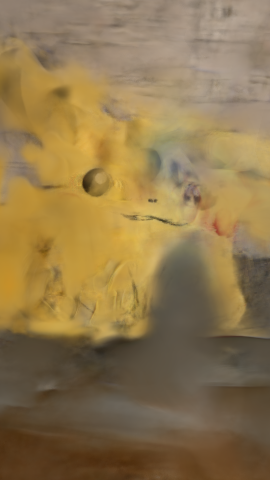}
    \end{minipage}} 
    \subfigure[CF-NeRF]{
    \begin{minipage}[b]{0.13\linewidth}
        \includegraphics[width=1\linewidth]{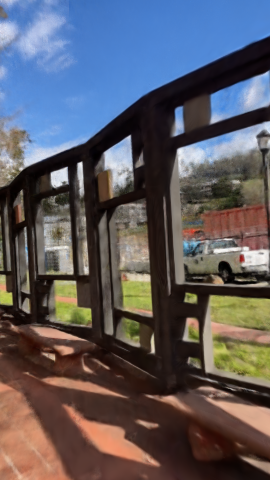}
        \includegraphics[width=1\linewidth]{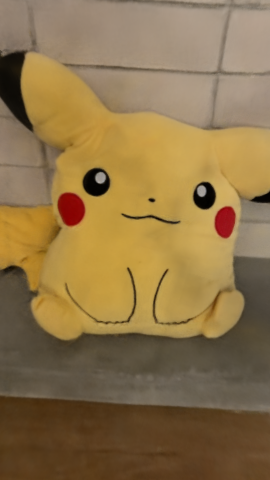}
    \end{minipage}}
    
	\vspace{-1.0em}
	
	\caption{
		\textbf{Qualitative results on NeRFBuster.} 
		We select two sequences from NeRFBuster \cite{warburg2023nerfbusters} and render novel views to compare the quality of camera parameters. Our method CF-NeRF generate high-quality images, while results of NeRFmm \cite{wang2021nerf}, SiRENmm \cite{ventusff2021}, BARF \cite{lin2021barf}, GARF \cite{chng2022garf} and L2G-NeRF \cite{chen2023local} contain lots of noise.
	}
	\label{fig:viz}
	
	\vspace{-1.5em}
    
\end{figure*}

\paragraph{Initialization}

Similar to incremental SfM, CF-NeRF requires initialize $\theta$, $\delta_{R_1}$, $\delta_{T_1}$, and $\delta_{K}$ before adding images to $E$. 
We select the first $N_{init}$ images $I_{init}$ from $I$ to optimise these parameters by Eq. \ref{eq:init_op} with $\xi_{init}$ iterations. 
Since the rotation between adjacent images is not large and NeRF is hard to estimate rotation \cite{lin2021barf,chng2022garf, chen2023local}, we do not estimate the rotation in the initialization to reduce the freedom. After initialization, we add $I_1$ to $E$ and keep $\theta$, $\delta_{R_1}$, $\delta_{T_1}$, and $\delta_{K}$ but discard other camera parameters. Note that, unlike the initialization in the previous section, the initialization here is data-specific, similar to the warm-up procedure.

\begin{align}
    \label{eq:init_op}
	{ \underset{\theta,\delta_T,\delta_K}{{\arg\min}\,}  \{ 
    \sum_{I_i \in I_{init}} 
    \sum_{p \in I_i}
    { \mathcal{L} ( \mathcal{R}(p|\theta, \delta_{T_i},\delta_K),I_i(p) ) \} } }
\end{align}

\paragraph{Implicit Localization}

After initialization, CF-NeRF estimates the camera parameter of the remaining images one by one and determines $\delta_{R_n}$ and $\delta_{T_n}$ for each new image $I_n$ by localization. Specifically, we first initialize $\delta_{R_n}$ and $\delta_{T_n}$ by $\delta_{R_{n-1}}$ and $\delta_{T_{n-1}}$, and then optimize them by minimizing  Eq. \ref{eq:local_op} with fixed $\theta$ through $\xi_{loc}$ iterations. The localization is similar to iNeRF \cite{yen2021inerf}, but CF-NeRF does not have a pre-trained $\mathcal{F}$.

\begin{align}
    \label{eq:local_op}
	{ \underset{\delta_{R_n},\delta_{T_n}}{{\arg\min}\,}  \{ 
    \sum_{p \in I_{n}}
    { \mathcal{L} ( \mathcal(p|\delta_{R_n},\delta_{T_n}}),I_{n}(p) ) } \} 
\end{align}

\paragraph{Implicit Optimization}

Although implicit localization can roughly determine $\delta_{R_n}$ and $\delta_{T_n}$, it faces two problems: the observation from $I_n$ is not added to NeRF, and the localization does not take the multi-view consistency into account to reduce drift. Incremental SfM solves these problems using two separate steps: triangulation and BA, while CF-NeRF benefits from the volume rendering and deals with these problems together. 
However, it is time-consuming to optimize all images in $E$ every time a new image is added. Therefore, CF-NeRF splits optimization into implicit partial optimization and implicit global optimization.

Each time localizing a new image $I_n$, CF-NeRF performs implicit partial optimization. We select $I_n$ and previous $N_{part}-1$ images to construct the partial image set $I_{part}$, then optimizes them with $\xi_{part}$ iterations, as Eq.\ref{eq:part_op} shows.

\begin{align}
    \label{eq:part_op}
	{ \underset{\theta,\delta_R,\delta_T}{{\arg\min}\,}  \{ 
    \sum_{I_i \in I_{part}} 
    \sum_{p \in I_i}
    { \mathcal{L} ( \mathcal{R}(p|\theta, \delta_{R_i}, \delta_{T_i}),I_i(p) ) \} } }
\end{align}

When the number of images in $E$ can be evenly divided by $N_{glob}$, CF-NeRF employs implicit global optimization for $\theta$ and all images in $E$ to enhance the overall accuracy and reduce drifts with $\xi_{glob}$ iterations, as Eq. \ref{eq:glob_op} shows.

\begin{align}
    \label{eq:glob_op}
	{ \underset{\theta,\delta_R,\delta_T,\delta_K}{{\arg\min}\,}  \{ 
    \sum_{I_i \in I_E} 
    \sum_{p \in I_i}
    { \mathcal{L} ( \mathcal{R}(p|\theta, \delta_{R_i}\delta_{T_i},\delta_K),I_i(p) ) \} } }
\end{align}

\begin{table*}
    \scriptsize
    \begin{center}
        \caption{ 
        We quantitatively evaluated on 12 scenes from the NeRFBuster dataset \cite{warburg2023nerfbusters}, which is captured in the real world with complex trajectories, including rotations and forward-looking. 
        CF-NeRF achieves state-of-the-art in $\Delta R$, $\Delta T$, PSNR, and LPIPS compared to NeRFmm \cite{wang2021nerf}, SiRENmm \cite{ventusff2021}, BARF \cite{lin2021barf}, GARF \cite{chng2022garf}, L2G-NeRF \cite{chen2023local}. 
        }

        \vspace{-1.0 em}
        
        \label{tab:pose_results}
        \resizebox{\linewidth}{!}{\begin{tabular}{c|l|cccccccccccc}
                
                \hline\noalign{\smallskip}
                
                & & \multicolumn{12}{c}{NeRFBuster }  \\

                \hline

                & & aloe & art & car & century & 
                flowers & garbage & picnic & pikachu 
                & pipe & plant & roses & table \\

                \hline

                \multirow{6}{*}{$\Delta R \downarrow$}
                & NeRFmm & 
                159.9731 & 177.5907 & 129.5799 & 
                119.6261 & 106.9195 & 150.8233 & 
                154.7781 & 113.7001 & 164.8207 & 
                165.0298 & 102.2748 & 115.2991 \\
                & SiRENmm & 
                155.1513 & 177.3642 & 127.2673 & 
                 89.0172 & 103.8741 &  82.9375 & 
                 44.3671 &  25.3603 & 159.7565 & 
                114.0761 & 132.5378 &  93.2612 \\
                & BARF & 
                158.6689 &  59.1868 & 133.4527 & 
                101.6013 &  88.6842 &  88.7832 & 
                 69.7201 &  41.0302 &  64.5250 & 
                143.1975 & 133.7586 & 111.2884 \\
                & GARF & 
                125.9797 & 171.9169 & 153.5585 & 
                105.1868 & 106.0595 &  84.4992 & 
                 49.8503 &  32.7285 & 126.9602 & 
                156.6058 & 118.9748 & 164.6562 \\
                & L2G-NeRF &
                124.2368 &  24.3753 &  55.7291 &
                131.9683 &  96.3498 & 110.4782 &
                146.3119 & 116.8911 &  70.9131 &
                 70.7562 &  95.9982 & 116.4972 \\
                & CF-NeRF & 
                \textbf{12.1226} & \textbf{19.2496} & \textbf{17.5570} & 
                \textbf{9.6811} & \textbf{8.2556} & \textbf{9.7658} & 
                \textbf{12.6501} & \textbf{11.3067} & \textbf{19.9926} & 
                \textbf{4.8968} & \textbf{5.1229} & \textbf{4.5837} \\

                \hline

                \multirow{6}{*}{$\Delta T \downarrow$}
                & NeRFmm & 
                11.5935 & 15.0762 & 23.9514 & 
                24.5934 & 12.8753 & 16.3842 & 
                12.9675 & 25.6841 & 19.3563 & 
                23.6613 &  8.0367 & 13.3849) \\
                & SiRENmm & 
                11.4912 & 14.8720 & 27.7235 & 
                28.8582 & 15.4841 & 13.3099 & 
                 8.3607 & 15.8052 & 20.4572 & 
                31.2943 &  9.0498 & 13.8350 \\
                & BARF & 
                 9.6196 & 17.3299 & 36.0351 & 
                26.0549 & 15.3166 & 17.1936 & 
                 9.6679 & 18.7184 & 18.8936 & 
                31.8629 &  8.5266 & 17.5453 \\
                & GARF & 
                13.9100 & 17.1527 & 26.5184 & 
                26.0096 & 14.1459 & 16.1054 & 
                11.7021 & 17.4975 & 18.8366 & 
                32.3170 & 11.2654 & 15.0367 \\
                & L2G-NeRF &
                14.4012 & 13.4240 & 20.5634 &
                23.2650 &  7.4559 & 17.7167 &
                12.9408 & 32.4048 & 11.4012 &
                18.5012 & 10.7110 & 12.8061 \\
                & CF-NeRF & 
                \textbf{3.3788} & \textbf{2.2821} & \textbf{6.5452} & 
                \textbf{2.7383} & \textbf{2.7026} & \textbf{4.0535} & 
                \textbf{1.2833} & \textbf{4.0586} & \textbf{9.4491} & 
                \textbf{3.4346} & \textbf{1.1945} & \textbf{1.2127} \\

                \hline

                \multirow{6}{*}{PSNR$\uparrow$}
                & NeRFmm & 
                20.6912 & 16.8220 & 17.7504 & 
                16.0326 & 18.4377 & 17.7229 & 
                18.4300 & 25.3819 & 20.0978 & 
                21.1558 & 17.7735 & 14.1651 \\
                & SiRENmm & 
                22.7462 & 20.3890 & 22.0268 & 
				18.0252 & 19.4640 & 17.2283 & 
				21.5628 & 27.8706 & 20.9538 & 
				23.4980 & 16.7480 & 18.8135 \\
                & BARF & 
                22.4366 & 21.1947 & 16.7665 & 
				15.3436 & 17.8350 & 15.9065 & 
				19.1846 & 23.0386 & 19.9728 & 
				25.5135 & 13.6741 & 13.8227 \\
                & GARF & 
                19.0241 & 19.3556 & 15.4460 & 
				14.4117 & 16.2955 & 15.3383 & 
				15.4035 & 20.9663 & 18.5371 & 
				20.5600 & 13.1274 & 12.6677 \\
                & L2G-NeRF & 
				21.3398 & 19.8099 & 17.3255 & 
				16.6476 & 18.0016 & 13.6077 & 
				18.5268 & 22.4939 & 18.1787 & 
				19.0160 & 17.2614 & 15.5658 \\
                & CF-NeRF & 
                \textbf{26.9367} & \textbf{26.5293} & \textbf{22.4654} & 
				\textbf{21.7072} & \textbf{21.6950} & \textbf{22.4736} & 
				\textbf{22.5475} & \textbf{32.3661} & \textbf{22.2719} & 
				\textbf{25.7312} & \textbf{24.3918} & \textbf{26.8491} \\

                \hline

                \multirow{6}{*}{LPIPS$\downarrow$}
				& NeRFmm & 
				0.5560 & 0.4954 & 0.5991 & 
				0.5793 & 0.5778 & 0.5661 & 
				0.6113 & 0.3683 & 0.5614 & 
				0.4927 & 0.5371 & 0.6073 \\
				& SiRENmm & 
				0.4508 & 0.4034 & 0.4450 & 
				0.4785 & 0.5048 & 0.5193 & 
				0.5227 & 0.2883 & 0.5170 & 
				0.3256 & 0.5333 & 0.4659 \\
				& BARF & 
				0.3328 & 0.3511 & 0.5361 & 
				0.5394 & 0.5552 & 0.5480 & 
				0.5358 & 0.3440 & 0.5198 & 
				0.3217 & 0.6138 & 0.5913 \\
				& GARF & 
				0.5257 & 0.4055 & 0.5984 & 
				0.5845 & 0.6158 & 0.5931 & 
				0.6086 & 0.3987 & 0.5688 & 
				0.4345 & 0.6189 & 0.6356 \\
				& L2G-NeRF & 
				0.4620 & 0.4186 & 0.5409 & 
				0.5116 & 0.5466 & 0.6016 & 
				0.5530 & 0.4051 & 0.4741 & 
				0.3840 & 0.4788 & 0.5309 \\
				& CF-NeRF & 
				\textbf{0.1939} & \textbf{0.2316} & \textbf{0.3983} & 
				\textbf{0.3627} & \textbf{0.3983} & \textbf{0.3859} & 
				\textbf{0.4686} & \textbf{0.1679} & \textbf{0.4453} & 
				\textbf{0.2594} & \textbf{0.2831} & \textbf{0.3011} \\

				\hline

		\end{tabular}}
    \end{center}
    \vspace{-2.5em}
\end{table*}

\paragraph{Coarse-to-Fine } CF-NeRF uses a coarse-to-fine strategy to improve robustness. CF-NeRF first constructs a Gaussian pyramid with depth $d_G$, then recovers all parameters at a low-resolution image through the incremental pipeline. Finally, CF-NeRF directly performs implicit global optimization with a higher resolution in each scale of the Gaussian pyramid with $\xi_{G}$ iterations.

\paragraph{Loss Function} To improve robustness, we employ the Smooth-L1 loss function, as Eq. \ref{eq:loss} shows, where $gt$ represents the ground truth, $pr$ is the estimated value, and $\beta$ is the set to 1.0 by default.

\begin{equation}
    \label{eq:loss}
	\mathcal{L}(pr,gt) = 
    \begin{cases}
	    0.5*(gt-pr)^2/\beta & if |gt-pr|<\beta \\
        |gt-pr|-0.5*\beta & otherwise \\
	\end{cases}
\end{equation}

\section{Experiments}

\subsection{Dataset}

Instead of using the forward-looking dataset LLFF \cite{mildenhall2021nerf} as usually conducted in the previous studies, we use a real-world dataset from NeRFBuster \cite{warburg2023nerfbusters}, mainly rotating around an object, to evaluate our method.
We sample around 50 frames for each scene and resize all images to $480 \times $270.
We use COLMAP \cite{schonberger2016structure} to estimate ground truth (GT) camera parameters for all scenes. More details of the dataset can be found in the supplementary material.

\subsection{Implementation}

CF-NeRF is implemented using PyTorch. 
Similar to NeRFmm \cite{wang2021nerf}, CF-NeRF does not have hierarchical sampling and uses the coarse network, which has eight layers and the dimension of the hidden layers is set to 128. 
And we use the sine activation function instead of the ReLU, as SiRENmm \cite{ventusff2021} is more robust than NeRFmm.
We utilize the Adam optimizer \cite{kingma2014adam} to optimize all learnable parameters. Specifically, we set the learning rate of $\theta$ to 0.001, which undergoes a decay of 0.9954 every 200 epochs. Similarly, the learning rate of $\delta$ is set to 0.001 and undergoes a decay of 0.9000 every 2000 epochs.
Here we describe how to set the hyper-parameters in CF-NeRF.
We set $N_{init}$ and $N_{part}$ to 3 to meet the minimum requirements that can filter outliers based on MVG. 
To balance drift and efficiency, we set $N_{glob}$ to 5. 
Considering the input image resolution, we set $d_G$ to 3 to reconstruct all parameters by coarse-to-fine strategy.
The most important parameter in CF-NeRF is iteration, which is the epoch number for each image. 
During initialization, we set $\xi_{init}$ to 3000 to guarantee that $\theta$ and $\delta$ can be correctly initialized with fewer images.
Subsequently, during the incremental training, we maintain a consistent value of $\xi$, setting $\xi=\xi_{loc}=\xi_{part}=\xi_{glob}=\xi_G$ to 900, thus reconstructing the scene from images one by one.
Throughout all our experiments, we use the NVIDIA RTX3090.

\begin{table*}
    \scriptsize
    \begin{center}
        \caption{ \textbf{ Ablation experiments.} We compare the accuracy of camera parameter estimation of CF-NeRF under different hyper-parameter settings, including the iteration $\xi$, the global optimization frequency $N_{glob}$ and the coarse-to-fine strategy, where $C$ means the coarse stage and $F$ means the fine stage.
        }
        \vspace{-1.0 em}
        
        \label{tab:abla}
        \resizebox{\linewidth}{!}{\begin{tabular}{c|l|cccccccccccc}
                
                \hline\noalign{\smallskip}
                
                & & \multicolumn{12}{c}{NeRFBuster} \\

                \hline

                & & aloe & art & car & century & 
                flowers & garbage & picnic & pikachu 
                & pipe & plant & roses & table \\

                \hline

                \multirow{4}{*}{$\Delta R \downarrow$}
                & $F,\xi=600,N_{glob}=10$ & 
                17.8029 & 24.3389 &  17.1692 & 
                11.5924 & 11.6163 &   9.1240 & 
                14.6452 & 13.0037 &  19.0749 & 
                 5.3354 &  5.9091 &   6.4731 \\
                & $F,\xi=900,N_{glob}=10$ & 
                14.8730 & 22.8142 &  17.8879 & 
                11.1201 & 10.4707 &   \textbf{8.6973} & 
                \textbf{11.4209} & 12.0625 & 18.4305 & 
                 4.7303 &  5.8538 &   6.8481 \\
                & $C,\xi=900,N_{glob}=5$ & 
                12.4862 & \textbf{19.1647} & 17.4755 & 
                 9.7177 &  8.4555 &  9.6460 & 
                12.3162 & 10.9802 & 19.9855 & 
                5.1579 & 5.5133 & 5.2821 \\
                & $F,\xi=900,N_{glob}=5$ & 
                \textbf{12.1226} & 19.2496 & 17.5570 & 
                \textbf{9.6811} & \textbf{8.2556} & 9.7658 & 
                12.6501 & \textbf{11.3067} & 19.9926 & 
                \textbf{4.8968} & \textbf{5.1229} & \textbf{4.5837} \\

                \hline

                \multirow{4}{*}{$\Delta T \downarrow$}
                & $F,\xi=600,N_{glob}=10$ & 
                4.5457 & 5.9307 & 7.5697 & 
                2.9652 & 3.6234 & 4.5340 & 
                2.6677 & 5.3105 & 9.3544 & 
                4.4384 & 1.3324 & 1.9535 \\
                & $F,\xi=900,N_{glob}=10$ & 
                3.9111 & 6.1190 & 7.3752 & 
                3.6834 & 3.3956 & 4.3080 & 
                3.4918 & \textbf{3.2682} & \textbf{8.4666} & 
                3.6109 & 1.3013 & 2.3973 \\
                & $C,\xi=900,N_{glob}=5$ & 
                3.4681 & 2.2770 & 6.6250 & 
                2.8224 & 2.7405 & 4.1085 & 
                1.2886 & 4.2462 & 9.6998 & 
                3.5309 & 1.2182 & 1.2232 \\
                & $F,\xi=900,N_{glob}=5$ & 
                \textbf{3.3788} & \textbf{2.2821} & \textbf{6.5452} & 
                \textbf{2.7383} & \textbf{2.7026} & \textbf{4.0535} & 
                \textbf{1.2833} & 4.0586 & 9.4491 & 
                \textbf{3.4346} & \textbf{1.1945} & \textbf{1.2127} \\

                \hline

        \end{tabular}}
    \end{center}
    \vspace{-2.5 em}
\end{table*}

\subsection{Evaluation}

To demonstrate the performance of the proposed method, we conduct a comprehensive comparison between CF-NeRF and several state-of-the-art models,  including NeRFmm \cite{wang2021nerf} SiRENmm \cite{ventusff2021}, BARF \cite{lin2021barf}, GARF \cite{chng2022garf}, and L2G-NeRF \cite{chen2023local}. We use all images for camera parameter estimation without employing a train/test split.
To evaluate the quality of the camera parameters, we calculate the average translation error $\Delta T$ and the average rotation error $\Delta R$ by aligning the estimated camera parameters $\delta_R$ and $\delta_T$ with COLMAP using a similarity transformation Sim(3) \cite{lin2021barf}. 
It is worth noting that $\delta_T$ represents a relative translation error rather than an absolute measurement, as COLMAP can not reconstruct an absolute scale of the scene.
We further evaluate the estimated camera parameters through a novel view synthesis by PSNR and LPIPS. 
To ensure a fair comparison and avoid the influence of varying network backbones across different methods, we uniformly use the NerfAcc \cite{li2022nerfacc}, where we select one image for testing in every eight images and the remaining is for training.

\subsection{Results}

We performed qualitative and quantitative evaluations of these methods on 12 scenes of the NeRFBuster \cite{warburg2023nerfbusters} dataset. Notably, BARF \cite{lin2021barf}, GARF \cite{chng2022garf}, and L2G-NeRF \cite{chen2023local} require manual setting the focal length. In contrast, NeRFmm \cite{wang2021nerf}, SiRENmm \cite{ventusff2021}, and CF-NeRF have the ability to estimate the focal length.

Table \ref{tab:pose_results} shows the results of qualitative experiments.
Our method obtains the highest accuracy camera parameters, while all other methods fail outright. 
It is important to understand that $\Delta R$ and $\Delta T$ are calculated by aligning the camera positions with Sim(3) and that a slight difference in camera position can lead to huge errors.
The rotation error $\Delta R$ of our method CF-NeRF is roughly around $10^{\circ}$, while the other methods are around $100^{\circ}$. 
Moreover, the translation error $\delta_T$ of CF-NeRF is approximately about 4, while all other methods are around 15.
Although NeRFmm, SiRENmm, BARF, GARF, and L2G-NeRF claim high accuracy on forward-looking scenes from scratch, they are unsuitable for scenes with rotation and are prone to be trapped in a local minimum. 
In contrast, CF-NeRF recovers the camera parameters sequentially and can effectively handle image sequences with complex trajectories.
Furthermore, SiRENmm outperforms NeRFmm in camera parameter estimation, which is why CF-NeRF uses the sine activate function.

Table \ref{tab:pose_results} also shows the quality of the novel view synthesis, which serves as an additional evaluation criterion for the quality of camera parameters.
CF-NeRF achieves state-of-the-art results on PSNR and LPIPS, whereas other methods struggle to generate high-quality results across multiple novel views. 
Interestingly, the reconstruction results of other methods appear reasonable compared to their poor camera parameters, mainly due to the high over-fitting ability of NeRF and the fact that partial camera parameters are correctly reconstructed.
We further visualize the rendering results of three scenes from different methods in Figure \ref{fig:com} and Figure \ref{fig:viz}. CF-NeRF can generate high-quality results, while other methods have lots of noise in their results due to their inability to provide accurate camera parameters for all images.
More visualization results are in supplementary.

\subsection{Ablation Experiments}

To understand the influence of hyper-parameters in CF-NeRF, we conduct several ablation experiments on the iteration $\xi$, the global optimization frequency $N_{glob}$, and the coarse-to-fine strategy. These experiments are performed on the NeRFBuster dataset \cite{warburg2023nerfbusters}, and results are presented in Table \ref{tab:abla}.

\paragraph{The iteration $\xi$}

The iteration $\xi$ is the most important hyper-parameter in our method, determining how many times to optimize the camera parameter for each image. We compare two configurations: $F,\xi=600,N_{glob}=10$ and $F,\xi=900,N_{glob}=10$. Table \ref{tab:abla} reveals that increasing $\xi$ improves the final results for almost all scenes. This observation aligns with NeRF \cite{mildenhall2021nerf} and iNeRF \cite{yen2021inerf}, where NeRF requires a large number of iterations to converge, and iNeRF enhances the quality of camera parameters through more iterations.

\paragraph{The global optimization frequency  $N_{glob}$}

To mitigate drift while maintaining efficiency, CF-NeRF employs the implicit global optimization when every $N_{glob}$ image is added $E$.
We conduct two experiments $F,\xi=900,N_{glob}=10$ and $F,\xi=900,N_{glob}=5$ to find out the influence of $N_{glob}$. 
As highlighted in Table \ref{tab:abla}, reducing $N_{glob}$ yields improved final results, which can be attributed to the fact that global optimization ensures global consistency to avoid NeRF trap into a local minimum.

\paragraph{The coarse-to-fine stragety}

CF-NeRF adopts a coarse-to-fine strategy to avoid directly estimating camera parameters on high-resolution images, where the fine stage refines initial results from the coarse stage.
We conduct two experiments $C,\xi=900,N_{glob}=5$ and $F,\xi=900,N_{glob}=5$ , where $C$ means the coarse stage and $F$ means the fine stage.
Results in Table \ref{tab:abla} demonstrate that the fine stage outperforms the coarse stage across almost all scenes. 
Adopting the coarse-to-fine strategy facilitates the training process of CF-NeRF, as the pixel gradient is smoother at the coarse stage and has less RGB information to learn.

\subsection{Limitation}

Although CF-NeRF achieves state-of-the-art results in camera parameter estimation, surpassing other NeRF-based methods, there are still some gaps between CF-NeRF and COLMAP \cite{schonberger2016structure}, and the accuracy can be further improved through the adjustment of the sample space \cite{wang2023f2} or the utilization of a more robust function \cite{sabour2023robustnerf}.

\section{Conclusion}

This paper presents CF-NeRF, a novel end-to-end method that does not require prior camera parameters to deal with image sequences with complex trajectories. 
Following the pipeline of incremental SfM, CF-NeRF contains three major sub-modules: initialization, implicit localization, and implicit optimization. 
Experiments on the NeRFBuster dataset demonstrate that CF-NeRF achieves state-of-the-art results, while NeRFmm, SiRENmm, BARF, GARF, and L2G-NeRF only work for forward-looking scenes and get trapped in the local minimum on the NeRFBuster dataset.
More importantly, CF-NeRF highlights the unlimited potential of NeRF and differential volume rendering, showing that NeRF has impressive reconstruction capabilities and can also be used to estimate camera parameters in complex trajectories.

\paragraph{Acknowledgments}

The research was supported in part by a RGC RIF grant under the contract R6021-20, RGC CRF grants under the contracts C7004-22G and C1029-22G, and RGC GRF grants under the contracts 16209120, 16200221, and 16207922. 
This research was also supported by the National Natural Science Foundation of China (No. 62302126), and the Shenzhen Science and Technology Program (No. RCBS20221008093125065, No. JCYJ20220818102414030).

\vspace{3.5em}
\begin{center}
    {\Huge \bf Supplementary material}
\end{center}
\setcounter{section}{0}

\section{How to use COLMAP}

COLMAP  \cite{schonberger2016structure} is a widely used 3D reconstruction software that can produce stable reconstruction results, providing the intrinsic and extrinsic camera parameters and the sparse 3D point clouds. 
In this paper, we use COLMAP to process all datasets and use these results as the ground truth (GT) to evaluate camera parameters estimated by different methods. 
However, we make some modifications to the default settings of COLMAP to align with the unique characteristics of the datasets.

\begin{enumerate}
\item We assume no distortions in images and use the pinhole camera model that only considers the focal length. 
\item We upsample the image twice during feature extraction to ensure that COLMAP can extract enough feature points with traditional feature extraction algorithms. 
\item We use the sequential matching method to match images, which is much faster than the exhaustive matching method and more suitable for the incremental structure from motion (SfM) for an image sequence.
\end{enumerate}

\section{ Details of the NeRFBuster dataset }

\begin{figure*}
	
	\includegraphics[width=1.0\linewidth]{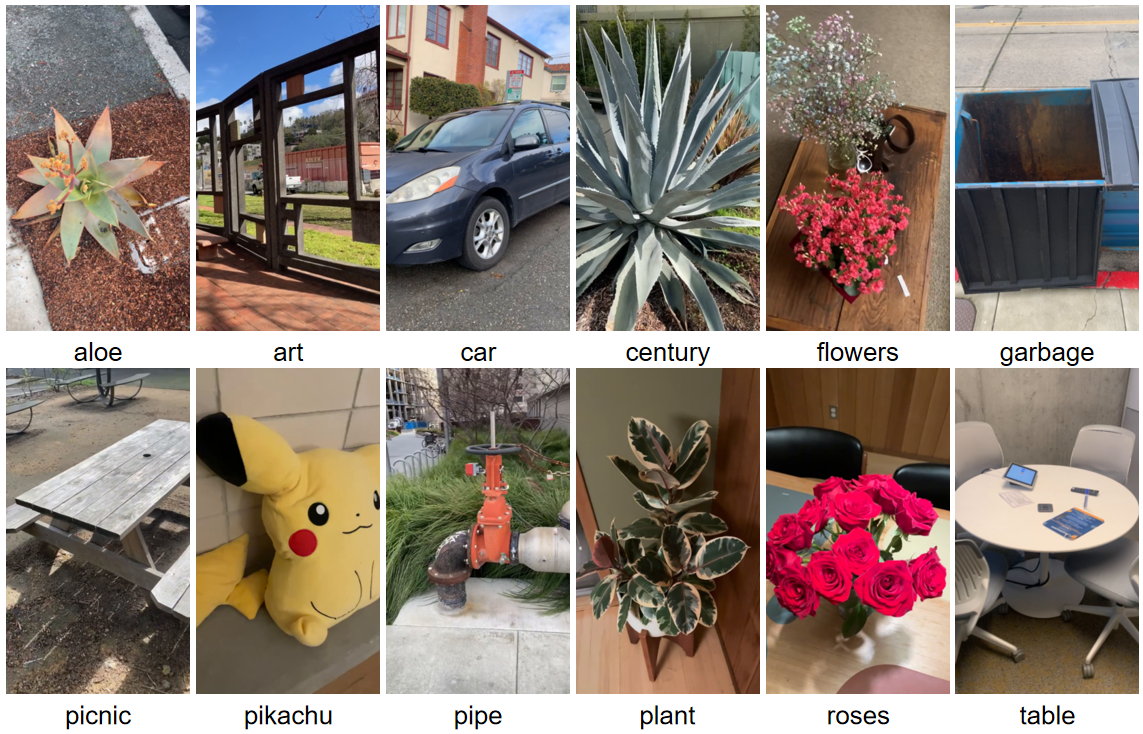}
	
	\caption{
		\textbf{12 scenes from NeRFBuster} We show one image from each scene in the NeRFBuster dataset \cite{warburg2023nerfbusters}, which contains various objects from indoor or outdoor environments.
	}
	
	\label{fig:scenes}
	
\end{figure*}

\begin{figure*}
	
	\includegraphics[width=1.0\linewidth]{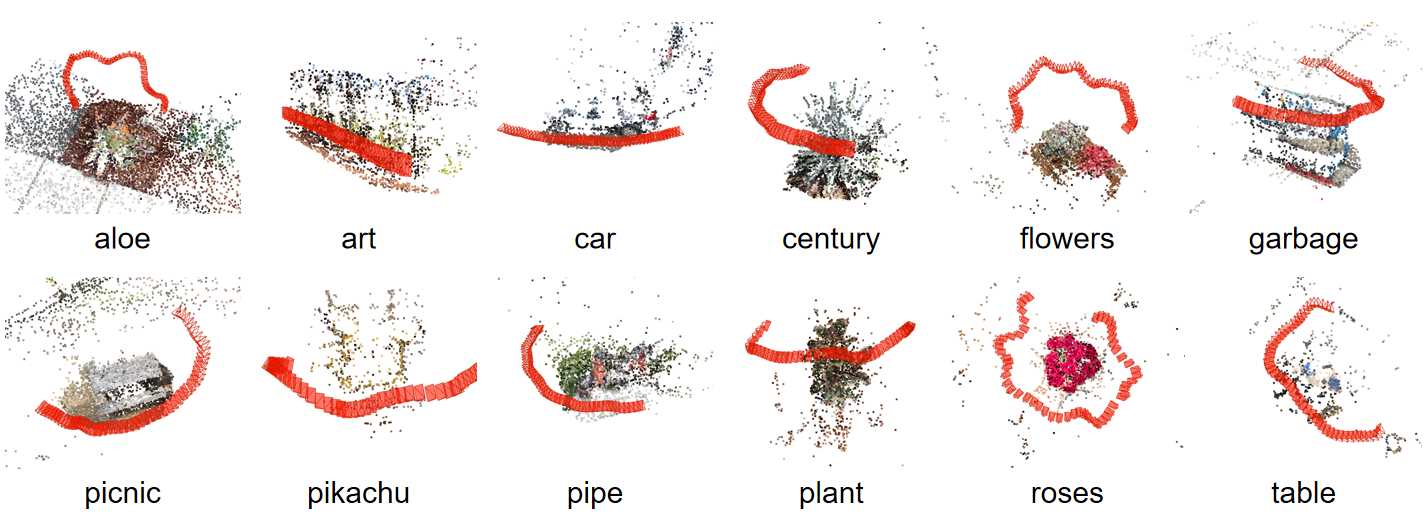}
	
	\caption{
		\textbf{Sparse reconstruction results of the NeRFBuster} The NeRFBuster dataset \cite{warburg2023nerfbusters} is captured under complex trajectories and mainly rotates around objects.
	}
	
	\label{fig:poses}
	
\end{figure*}

In all experiments, we use the challenging NeRFBuster dataset \cite{warburg2023nerfbusters}, which is captured in the real world. This dataset contains 12 scenes, including aloe, art, car, century, flowers, garbage, picnic, pikachu, pipe, plant, flowers, and table, as Figure \ref{fig:scenes} shows.
Furthermore, we present camera parameters and sparse point clouds results obtained from COLMAP in Figure \ref{fig:poses}.
Unlike the LLFF dataset \cite{mildenhall2021nerf}, which is limited to forward-looking motion, the NeRFBuster dataset is more challenging for camera parameters estimation, because it is captured under complex movement and contains a variety of objects.

\section{How to use NeRFacc}

Instead of only evaluating the camera parameter by comparing the camera translation and the camera rotation, we also indirectly evaluate the camera parameter by novel view synthesis. 
Because we use all images during camera parameter estimation and different methods use different backbones,  we uniformly use NeRFAcc \cite{li2022nerfacc} to measure Peak Signal-to-Noise Ratio (PSNR) and Learned Perceptual Image Patch Similarity (LPIPS). 
For each scene in NeRFBuster, we follow these steps:

\begin{enumerate}
\item We align the estimated camera parameters to the GT camera parameters provided by COLMAP through a similarity transformation Sim(3) \cite{lin2021barf}.
\item We split the images in each scene into a training/split set, with one-eighth of the images as test images and the rest as training images.
\item We use the bounding box of sparse points provided by the COLMAP and utilize NeRFAcc to reconstruct the scene with the aligned camera parameters implicitly.
\item We train the model for 20000 epochs on the train split and evaluate PSNR and LPIPS on the test split. 
\end{enumerate}

During the training process, we employ the default settings from NeRFAcc and utilize the MipNeRF \cite{barron2021mip} as the backbone, which is motivated by the abundance of open-space scenes present in the NeRFBuster dataset.

\section{How to align camera parameters}

We follow BARF \cite{lin2021barf} to align two camera parameters through Sim(3), where only the camera translation is used.
However, we find this method is unreliable when all cameras lie in a line or the camera translation contains noise.
To overcome this problem and improve robustness, we add a virtual point $(0,0,1)$ in the camera coordinate of each image and use the camera parameter to transform it to the world coordinate, thus using the camera rotation during the alignment process.

\section{Visualization Results}

We qualitatively compare the camera parameters of our method CF-NeRF with state-of-the-art methods NeRFmm \cite{wang2021nerf}, SiRENmm \cite{ventusff2021}, BARF \cite{lin2021barf}, GARF \cite{chng2022garf}, and L2G-NeRF \cite{chen2023local} on the NeRFBuster dataset \cite{warburg2023nerfbusters}. 
Figure \ref{fig:com} shows the camera parameter estimation results of the different methods as well as the sparse points of COLMAP, and it is not difficult to notice that only our CF-NeRF generates correct results, while the other methods generate entirely wrong results.
For a full comparison, we also provide a video in the supplementary, using NeRFAcc to render the video according to the camera parameters of the different methods.

\begin{table} [H]
    \scriptsize
    \begin{center}
        \caption{ \textbf{ Quantitative evaluation results.} We evaluate our method on LLFF and Real360 from NeRF. BARF \cite{lin2021barf} needs a pre-calculated focal length from COLMAP. NeRFmm \cite{wang2021nerf} and CF-NeRF (ours) do not need prior camera parameters. }
        \vspace{-1em}
        \label{tab:pose_results}
        \resizebox{\linewidth}{!}{\begin{tabular}{l|cc|c|cc|c|cc|c}
                \hline\noalign{\smallskip}
                scene & \multicolumn{3}{|c|}{rotation$\downarrow$} & \multicolumn{3}{c|}{translation$\downarrow$} & \multicolumn{3}{c}{PSNR$\uparrow$} \\
                \hline
                & NF$-$  & ours & BARF  & NF$-$ & ours & BARF & NF$-$  & ours & BARF \\
                \hline

                fern & 
                1.2724 & \textbf{0.8669} & 0.2244  & 
                0.2469 & \textbf{0.2218} & 0.0811 & 
                23.71 & \textbf{26.38} & 27.19 \\
                flower & 
                \textbf{4.5220} & 10.7385 & 1.1812 & 
                0.6570 & \textbf{0.5587} & 0.0482 & 
                23.04 & \textbf{26.80} & 29.06  \\
                fortress & 
                174.296 & \textbf{3.0570} & 0.2691 &
                6.0263 & \textbf{1.3652} & 0.0972 & 
                9.62 & \textbf{27.19} & 26.92  \\
                horns & 
                4.3958 & \textbf{3.3078} & 0.2477 &
                \textbf{0.7677} & 1.1096 & 0.0644 &
                18.74 & \textbf{26.71} & 29.29  \\
                leaves & 
                \textbf{0.4504} & 11.2940 & 1.7657 & 
                \textbf{0.2216} & 0.8991  & 0.1823 & 
                18.94 & \textbf{20.88} & 20.12  \\
                orchids & 
                4.7278 & \textbf{0.1951} & 0.2309 & 
                0.2728 & \textbf{0.0703} & 0.0641 & 
                14.79 & \textbf{22.26} & 22.21  \\
                room &
                \textbf{3.1178} & 3.7610 & 0.2552 & 
                \textbf{0.3987} & 2.2731  & 0.1712 & 
                24.20 & \textbf{28.76} & 32.63   \\
                trex & 
                5.0095 & \textbf{0.6298} & 1.0239 & 
                \textbf{0.3608}  & 0.4072 & 0.3525 & 
                23.61 & \textbf{28.37} & 28.86   \\
                
                \hline

                vasedeck & 
                148.9928 & \textbf{6.0560} & 165.0392 & 
                18.6165 & \textbf{0.3143} & 12.2520 & 
                16.64 & \textbf{23.63} & 14.74   \\
                pinecone & 
                122.9956 & \textbf{0.4887} & 42.6464 & 
                29.5298 & \textbf{0.1387} & 8.2233 & 
                10.66 & \textbf{23.80} & 10.88   \\
                \hline
                
        \end{tabular}}
    \end{center}
\end{table}

\section{Experiments on LLFF}

Throughout our paper, we use the NeRFBuster, since it is a real-world dataset captured by mobile phones and contains quite complex trajectories, whereas the commonly used LLFF contains only forward views with simple trajectories.
However, to fully compare wtih other methods, we conduct experiments on LLFF as well.
As shown in Table \ref{tab:pose_results}, our method achieves competitive results to NeRFmm and BARF (with pre-calculated focal length) in the eight forward-looking scenes of LLFF.
However, for the two scenes with rotation in NeRF360, our method achieves much better results than NeRFmm and BARF.

\begin{figure*}
    \centering
    
    \subfigure[GT]{
    \begin{minipage}[b]{0.133\linewidth}
    \includegraphics[width=1\linewidth]{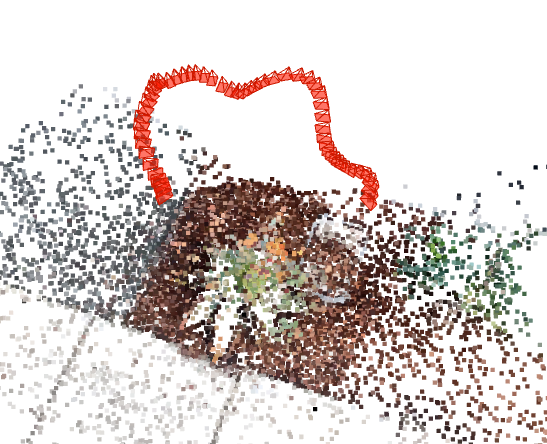} \vspace{-0.4em}
    \includegraphics[width=1\linewidth]{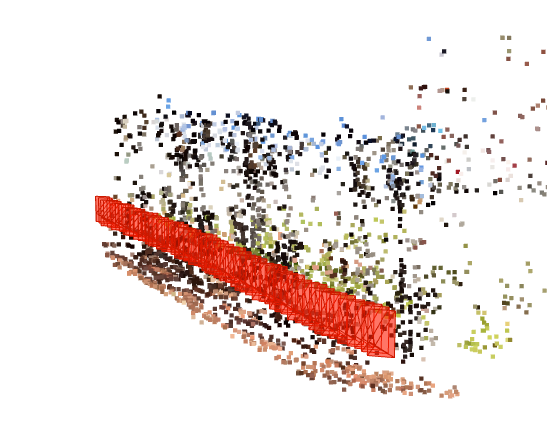} \vspace{-0.4em}
    \includegraphics[width=1\linewidth]{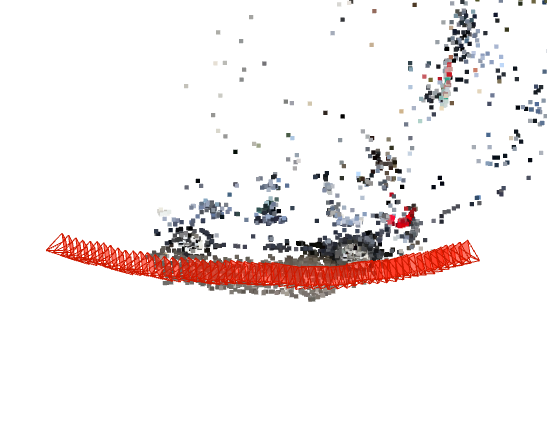} \vspace{-0.4em}
    \includegraphics[width=1\linewidth]{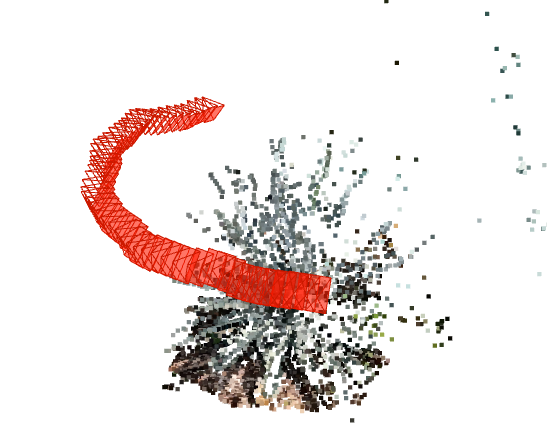} \vspace{-0.4em}
    \includegraphics[width=1\linewidth]{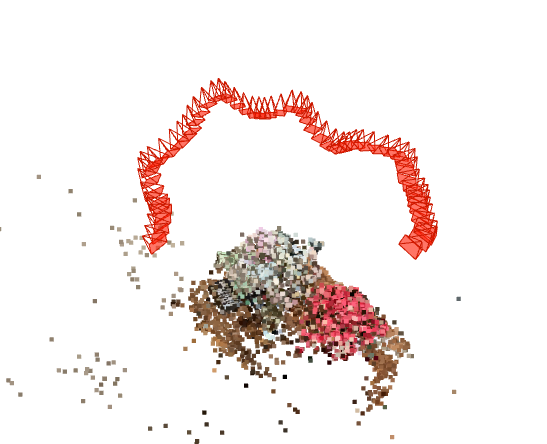} \vspace{-0.4em}
    \includegraphics[width=1\linewidth]{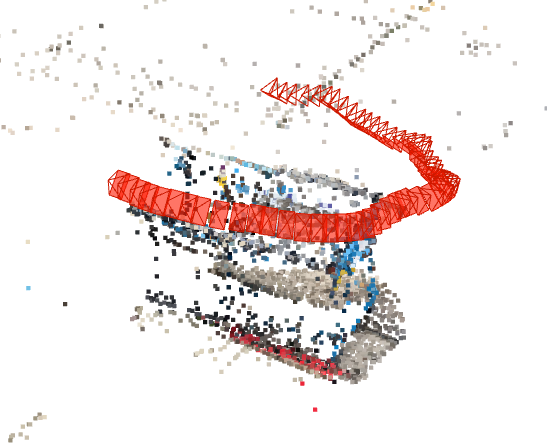} \vspace{-0.4em}
    \includegraphics[width=1\linewidth]{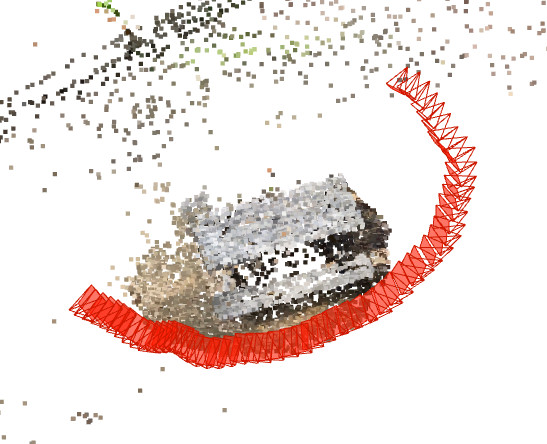} \vspace{-0.4em}
    \includegraphics[width=1\linewidth]{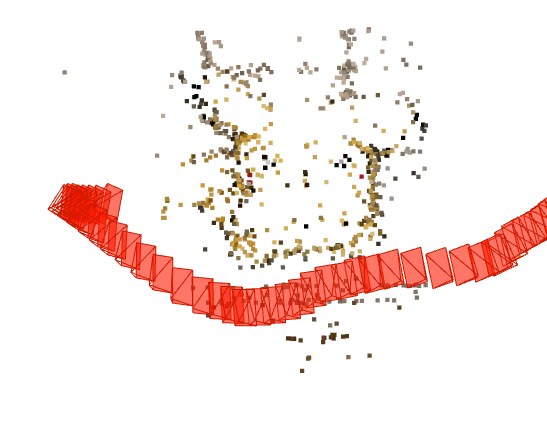} \vspace{-0.4em}
    \includegraphics[width=1\linewidth]{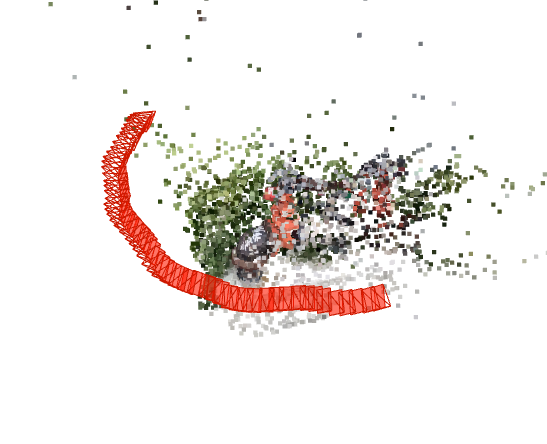} \vspace{-0.4em}
    \includegraphics[width=1\linewidth]{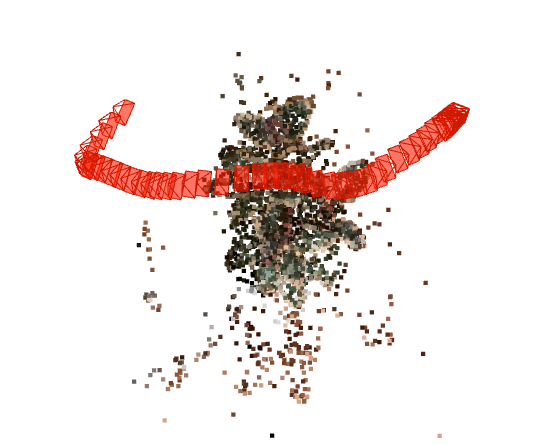} \vspace{-0.4em}
    \includegraphics[width=1\linewidth]{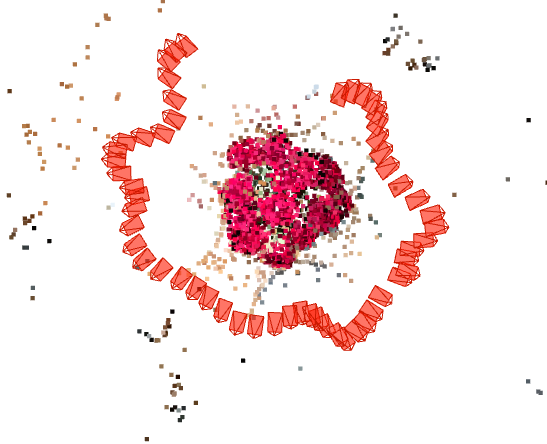} \vspace{-0.4em}
    \includegraphics[width=1\linewidth]{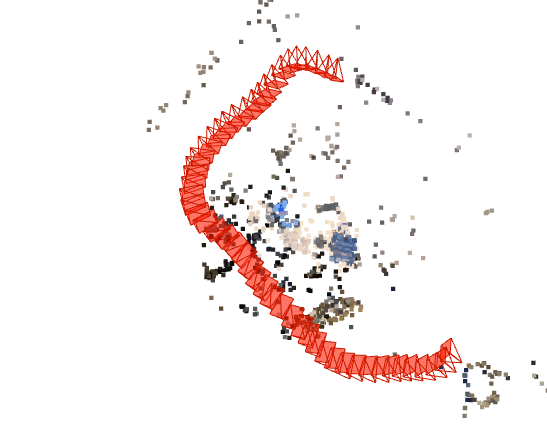}
    \end{minipage}}
    \subfigure[NeRFmm]{
    \begin{minipage}[b]{0.133\linewidth}
        \includegraphics[width=1\linewidth]{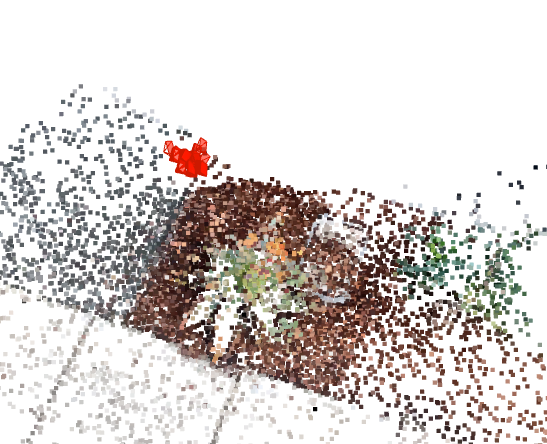} \vspace{-0.4em}
        \includegraphics[width=1\linewidth]{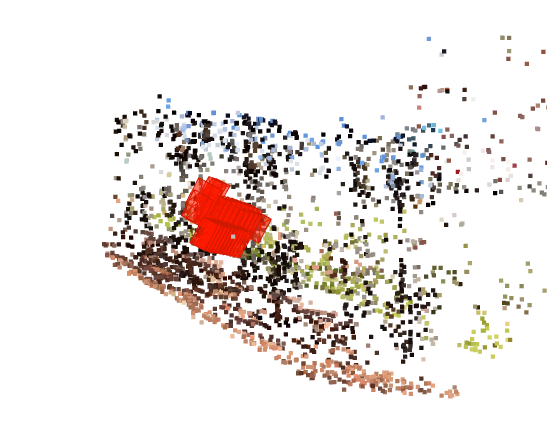} \vspace{-0.4em}
        \includegraphics[width=1\linewidth]{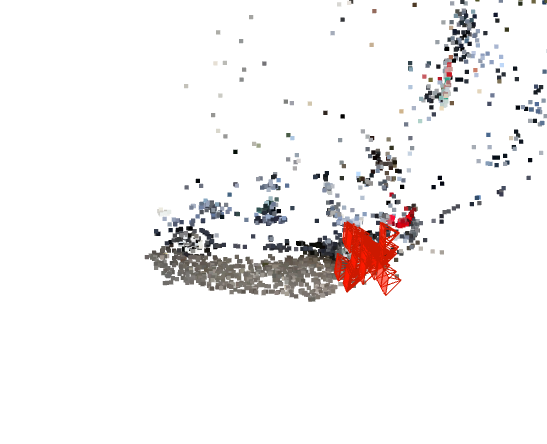} \vspace{-0.4em}
        \includegraphics[width=1\linewidth]{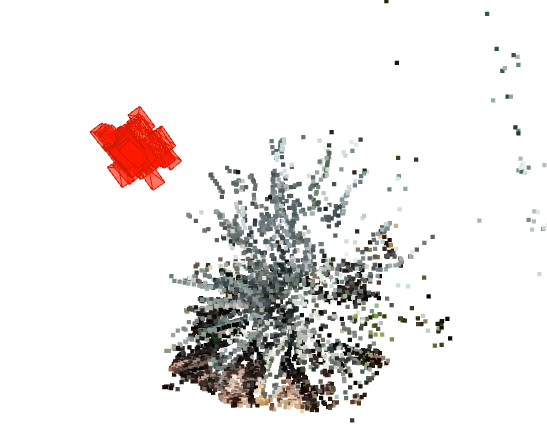} \vspace{-0.4em}
        \includegraphics[width=1\linewidth]{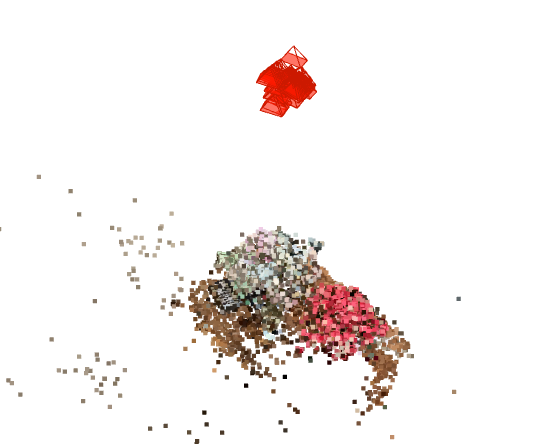} \vspace{-0.4em}
        \includegraphics[width=1\linewidth]{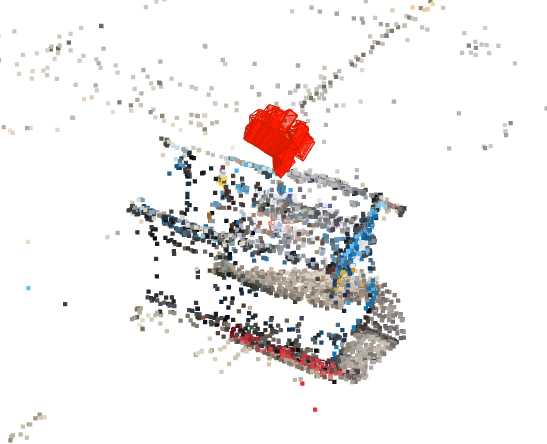} \vspace{-0.4em}
        \includegraphics[width=1\linewidth]{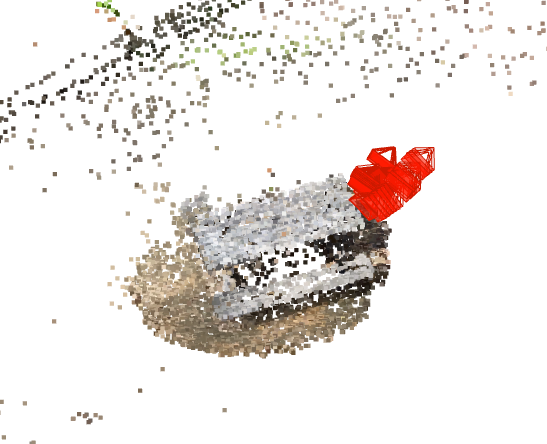} \vspace{-0.4em}
        \includegraphics[width=1\linewidth]{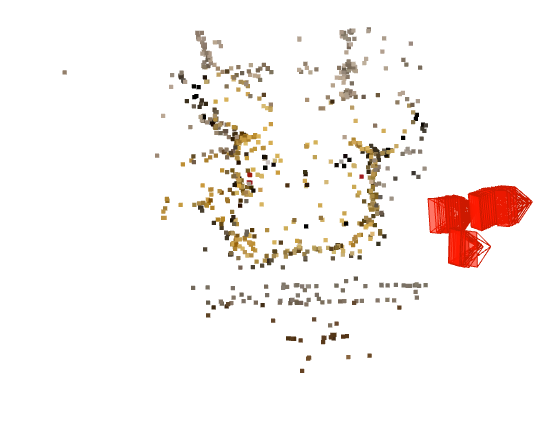} \vspace{-0.4em}
        \includegraphics[width=1\linewidth]{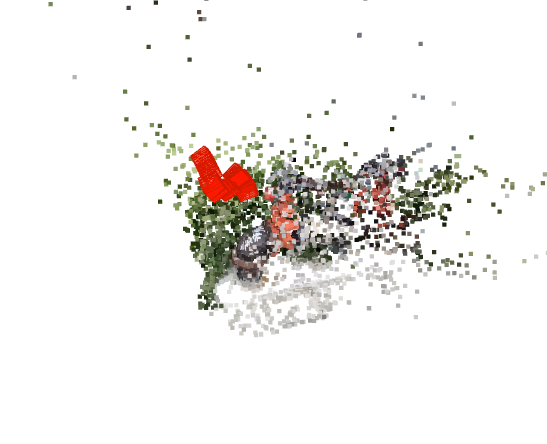} \vspace{-0.4em}
        \includegraphics[width=1\linewidth]{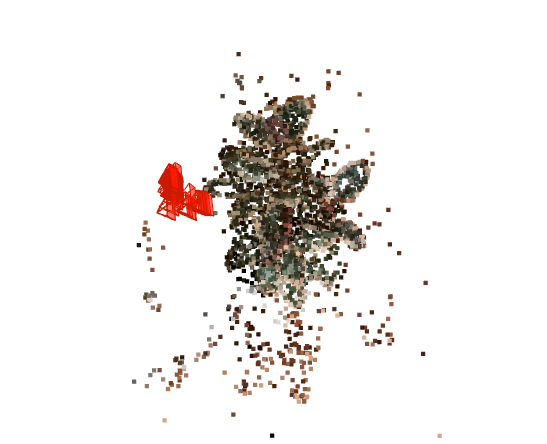} \vspace{-0.4em}
        \includegraphics[width=1\linewidth]{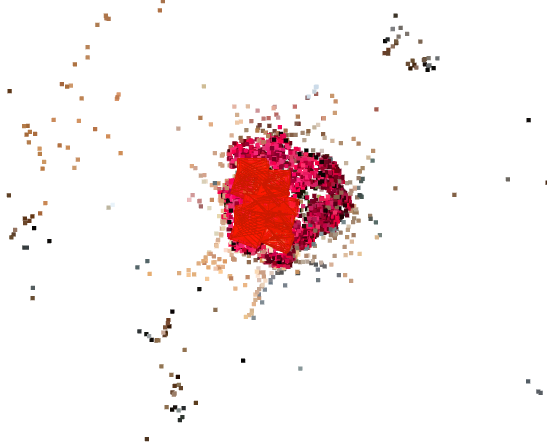} \vspace{-0.4em}
        \includegraphics[width=1\linewidth]{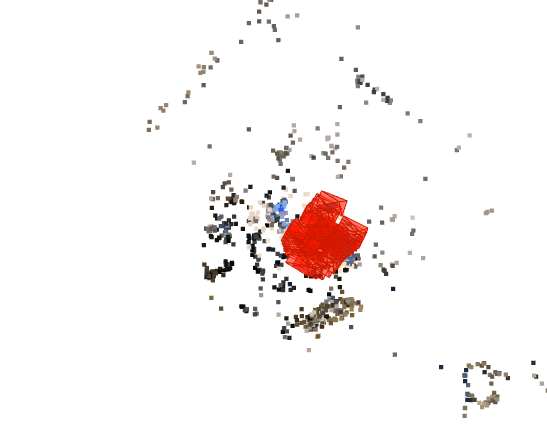}
    \end{minipage}}
    \subfigure[SiRENmm]{
    \begin{minipage}[b]{0.133\linewidth}
        \includegraphics[width=1\linewidth]{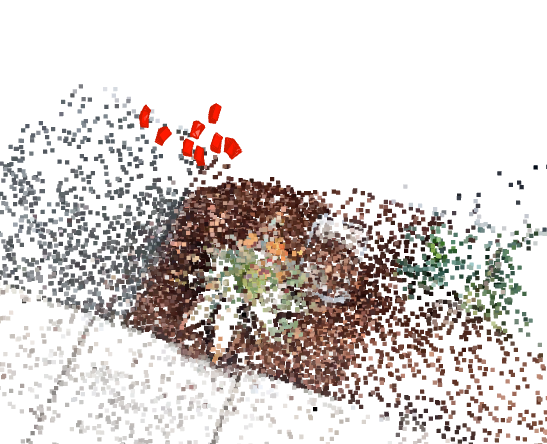} \vspace{-0.4em}
        \includegraphics[width=1\linewidth]{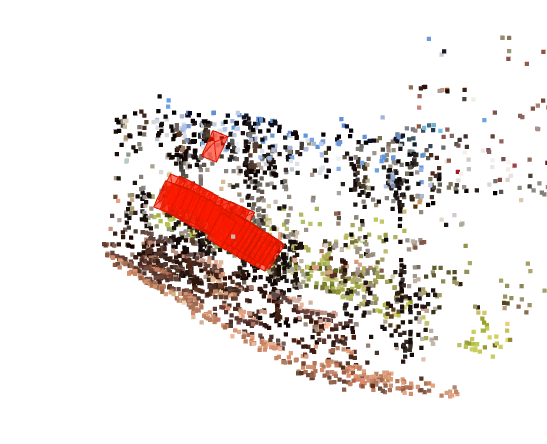} \vspace{-0.4em}
        \includegraphics[width=1\linewidth]{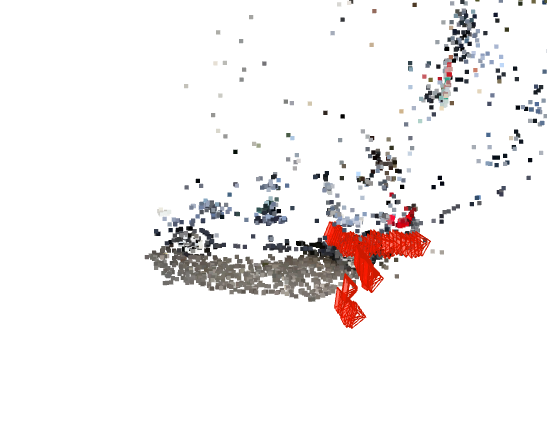} \vspace{-0.4em}
        \includegraphics[width=1\linewidth]{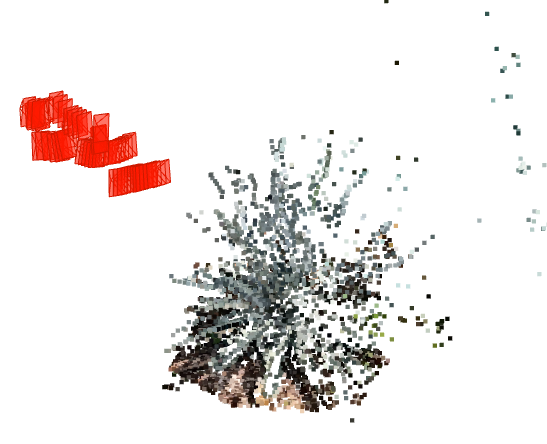} \vspace{-0.4em}
        \includegraphics[width=1\linewidth]{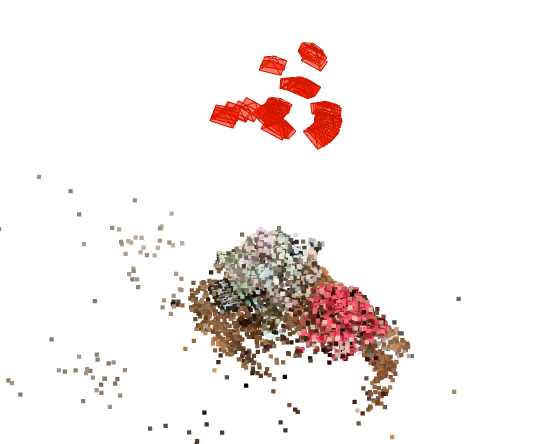} \vspace{-0.4em}
        \includegraphics[width=1\linewidth]{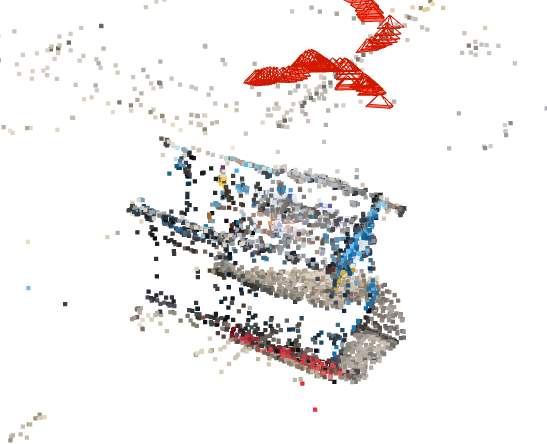} \vspace{-0.4em}
        \includegraphics[width=1\linewidth]{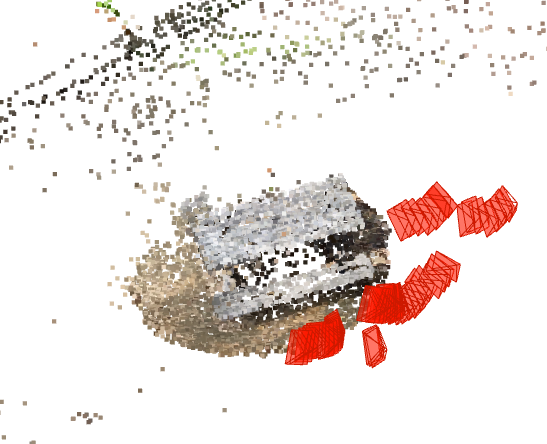} \vspace{-0.4em}
        \includegraphics[width=1\linewidth]{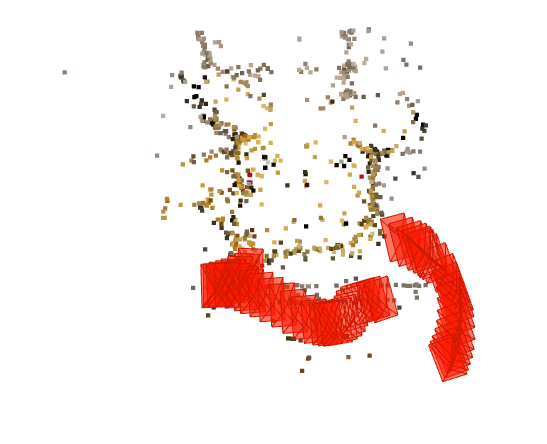} \vspace{-0.4em}
        \includegraphics[width=1\linewidth]{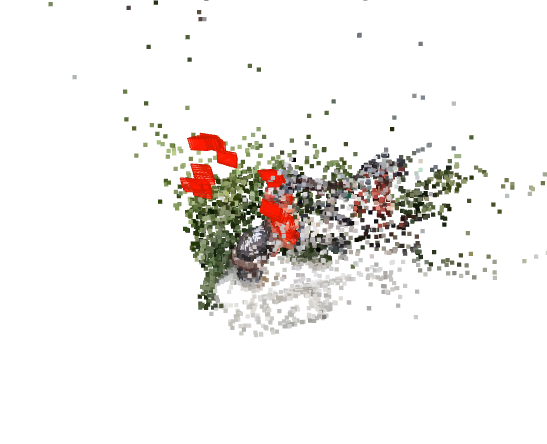} \vspace{-0.4em}
        \includegraphics[width=1\linewidth]{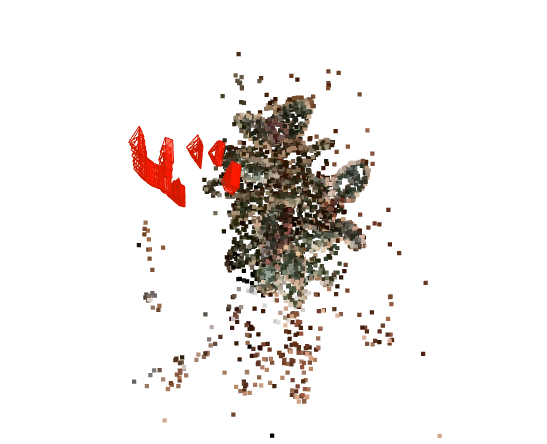} \vspace{-0.4em}
        \includegraphics[width=1\linewidth]{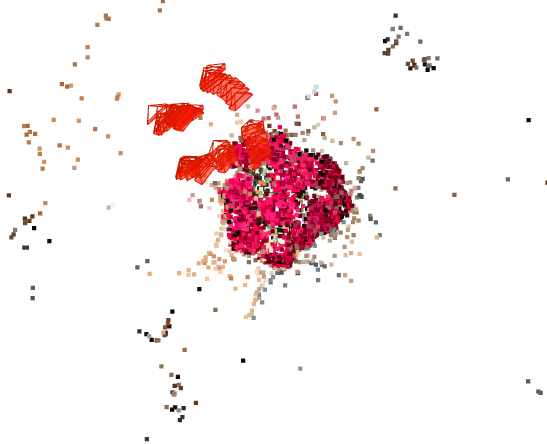} \vspace{-0.4em}
        \includegraphics[width=1\linewidth]{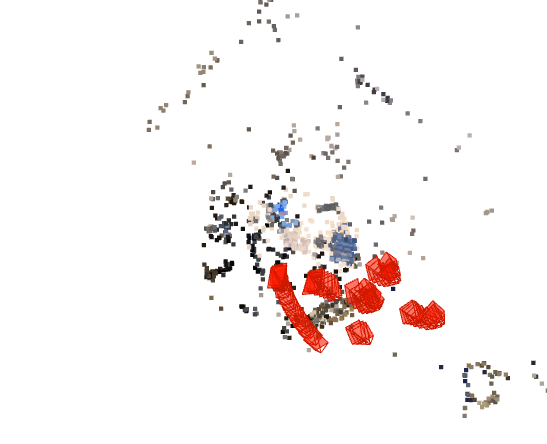}
    \end{minipage}}
    \subfigure[BARF]{
    \begin{minipage}[b]{0.133\linewidth}
        \includegraphics[width=1\linewidth]{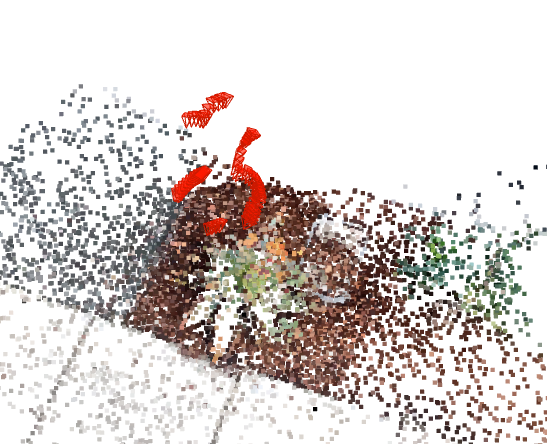} \vspace{-0.4em}
        \includegraphics[width=1\linewidth]{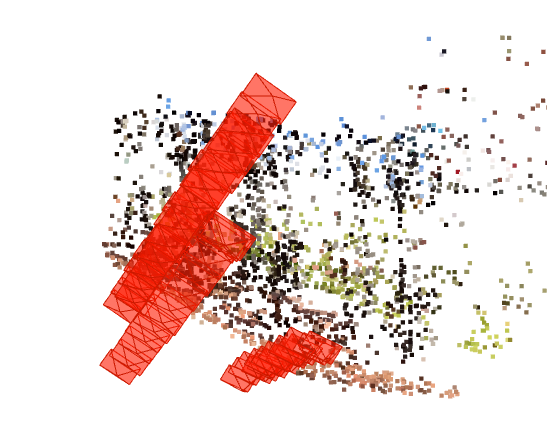} \vspace{-0.4em}
        \includegraphics[width=1\linewidth]{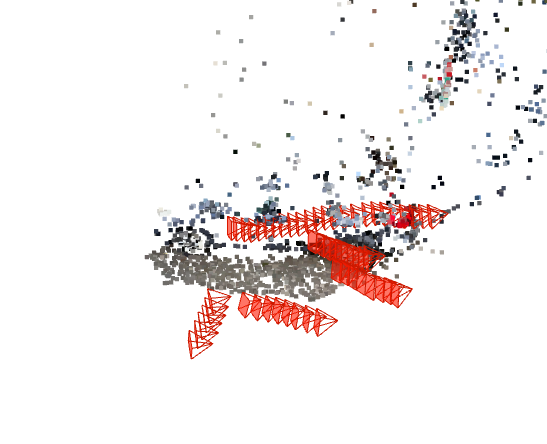} \vspace{-0.4em}
        \includegraphics[width=1\linewidth]{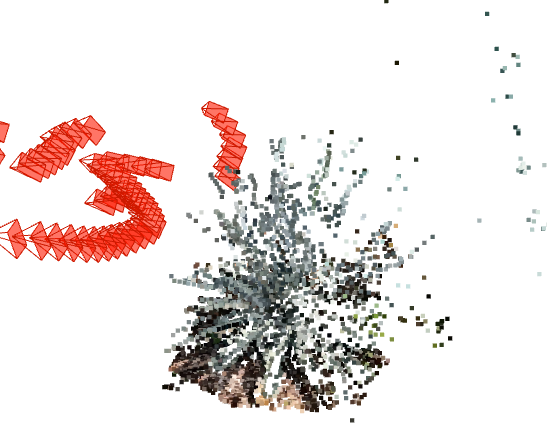} \vspace{-0.4em}
        \includegraphics[width=1\linewidth]{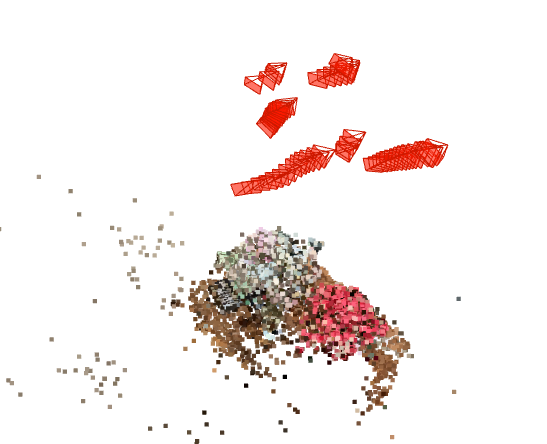} \vspace{-0.4em}
        \includegraphics[width=1\linewidth]{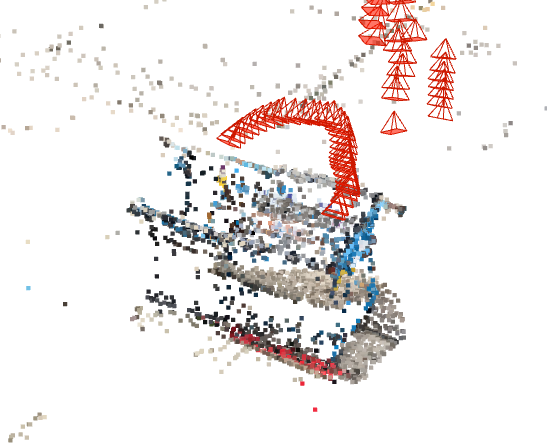} \vspace{-0.4em}
        \includegraphics[width=1\linewidth]{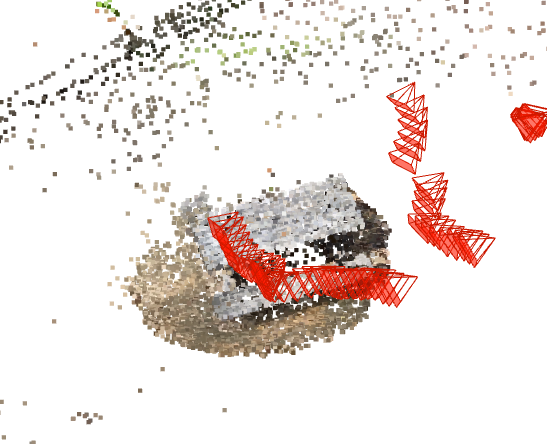} \vspace{-0.4em}
        \includegraphics[width=1\linewidth]{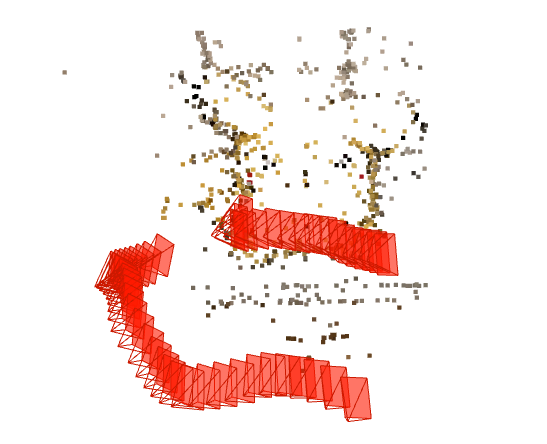} \vspace{-0.4em}
        \includegraphics[width=1\linewidth]{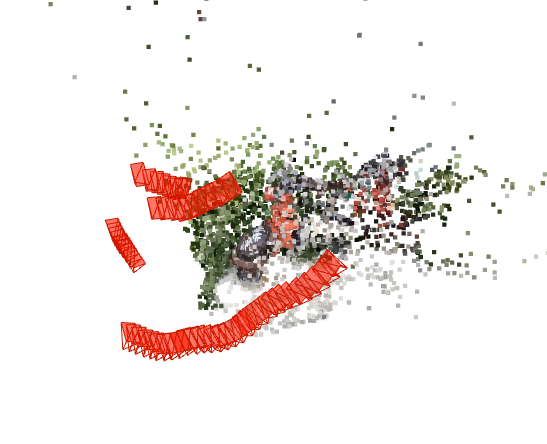} \vspace{-0.4em}
        \includegraphics[width=1\linewidth]{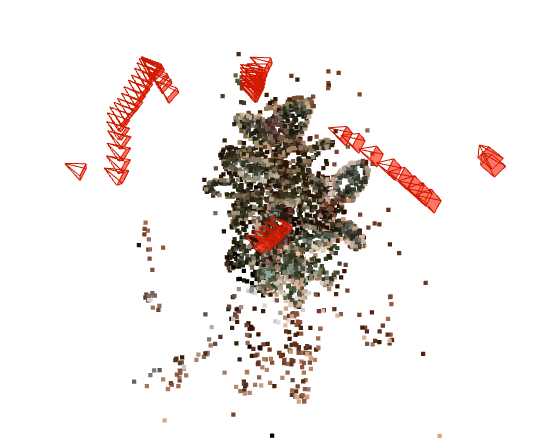} \vspace{-0.4em}
        \includegraphics[width=1\linewidth]{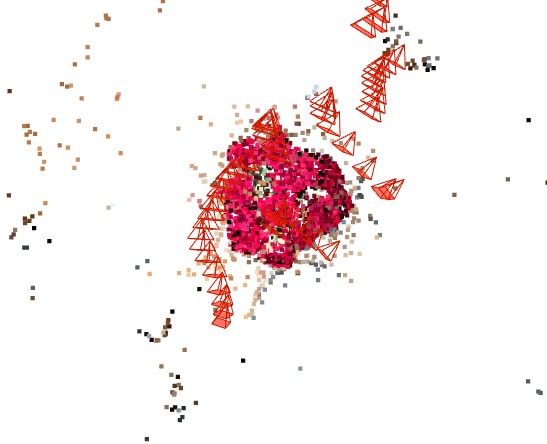} \vspace{-0.4em}
        \includegraphics[width=1\linewidth]{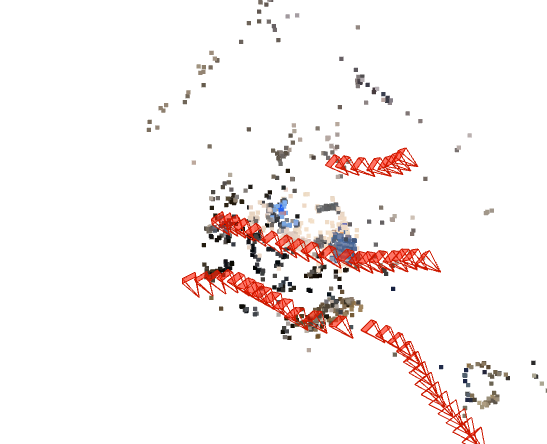}
    \end{minipage}}
    \subfigure[GARF]{
    \begin{minipage}[b]{0.133\linewidth}
        \includegraphics[width=1\linewidth]{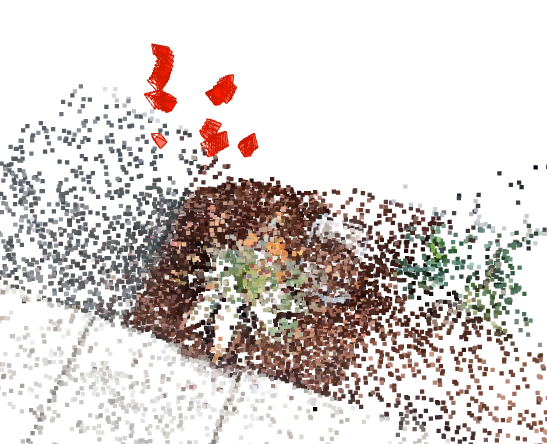} \vspace{-0.4em}
        \includegraphics[width=1\linewidth]{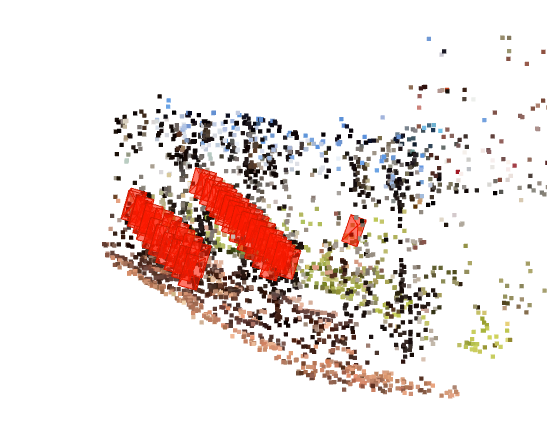} \vspace{-0.4em}
        \includegraphics[width=1\linewidth]{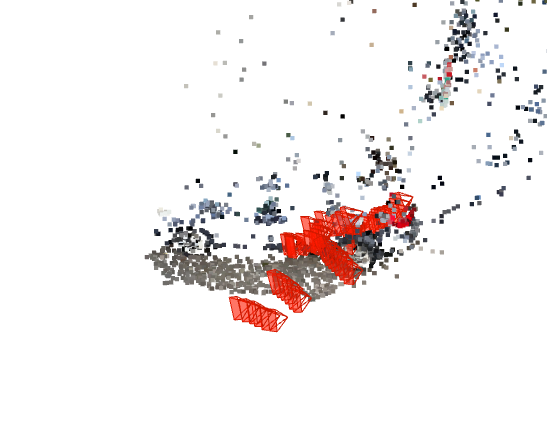} \vspace{-0.4em}
        \includegraphics[width=1\linewidth]{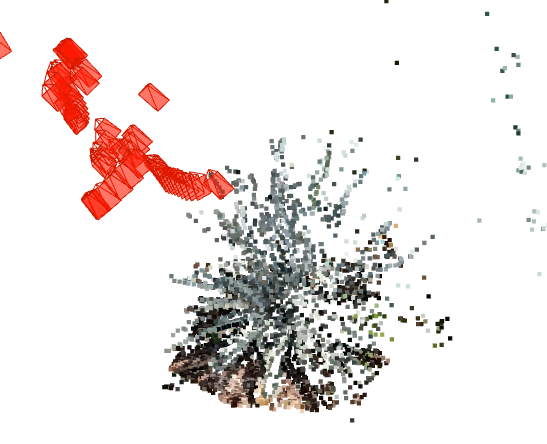} \vspace{-0.4em}
        \includegraphics[width=1\linewidth]{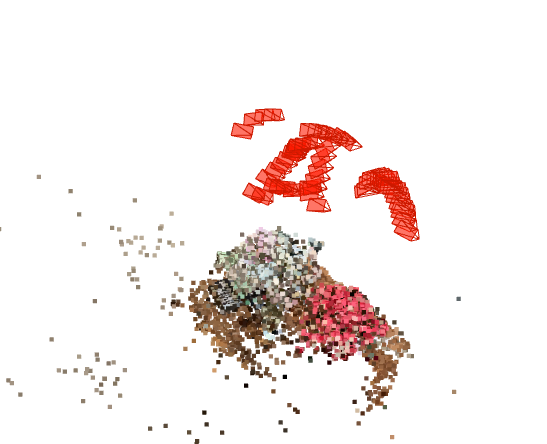} \vspace{-0.4em}
        \includegraphics[width=1\linewidth]{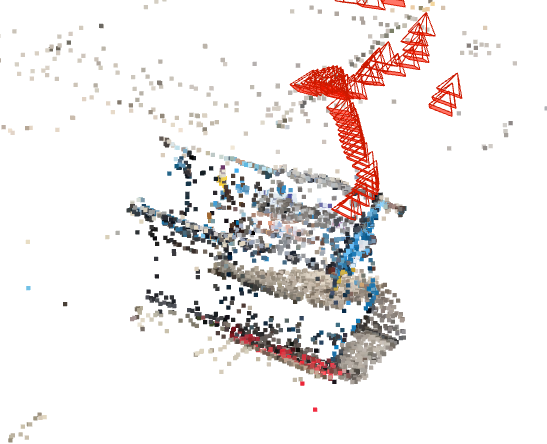} \vspace{-0.4em}
        \includegraphics[width=1\linewidth]{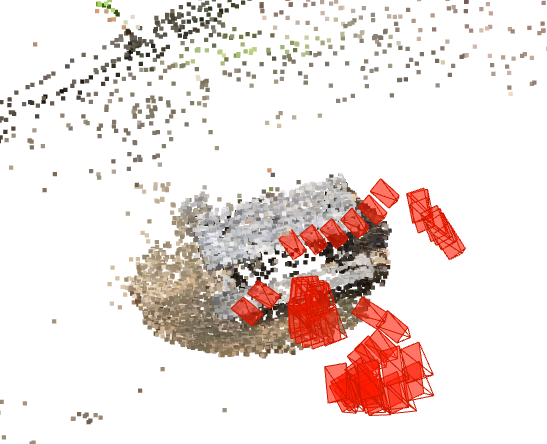} \vspace{-0.4em}
        \includegraphics[width=1\linewidth]{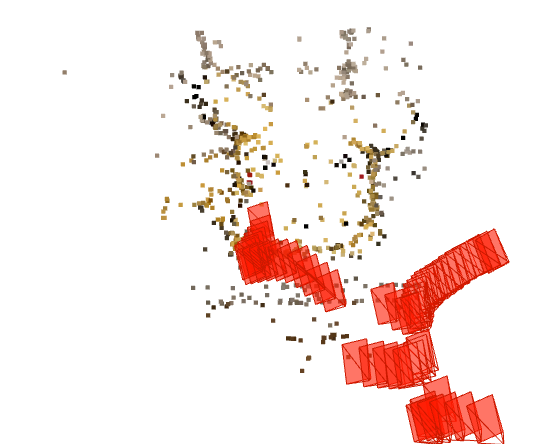} \vspace{-0.4em}
        \includegraphics[width=1\linewidth]{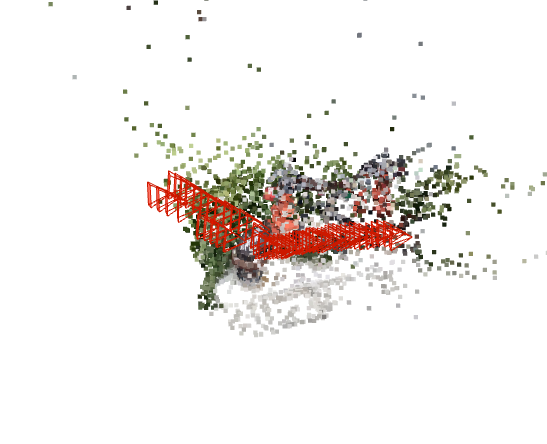} \vspace{-0.4em}
        \includegraphics[width=1\linewidth]{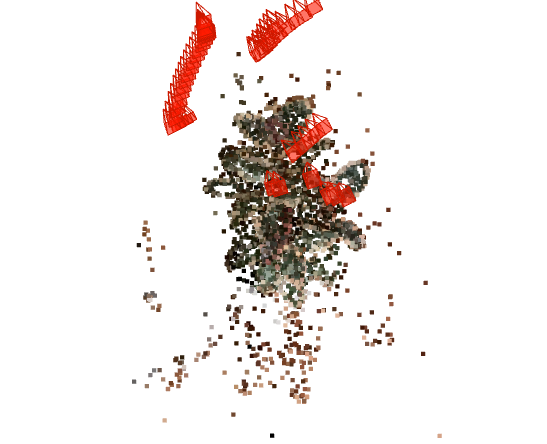} \vspace{-0.4em}
        \includegraphics[width=1\linewidth]{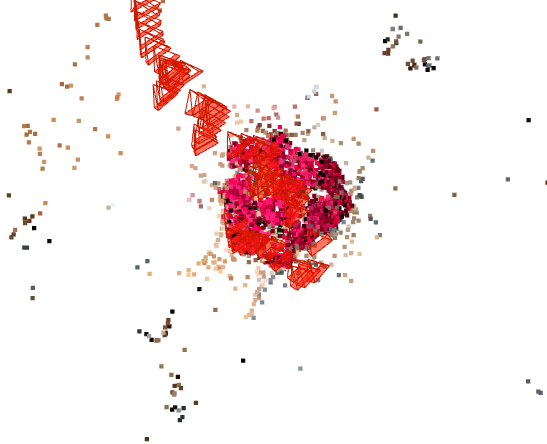} \vspace{-0.4em}
        \includegraphics[width=1\linewidth]{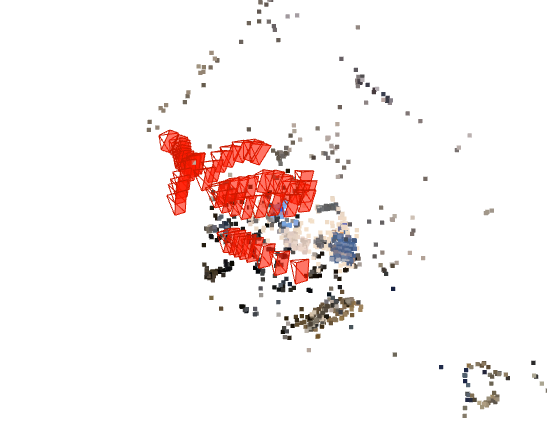}
    \end{minipage}}
    \subfigure[L2G-NeRF]{
    \begin{minipage}[b]{0.133\linewidth}
        \includegraphics[width=1\linewidth]{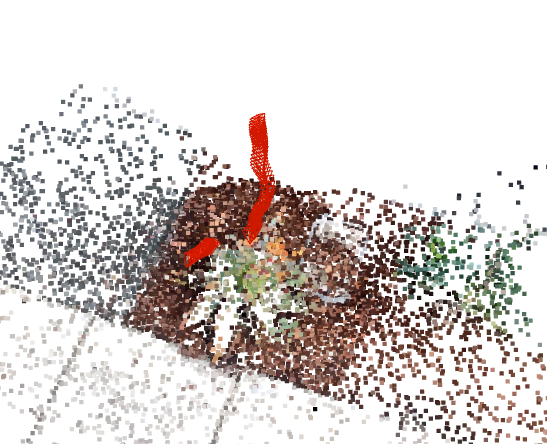} \vspace{-0.4em}
        \includegraphics[width=1\linewidth]{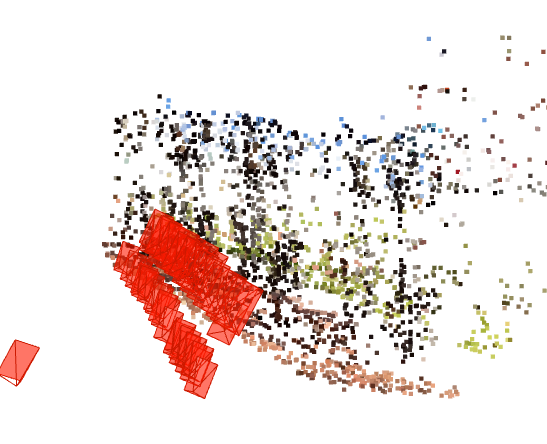} \vspace{-0.4em}
        \includegraphics[width=1\linewidth]{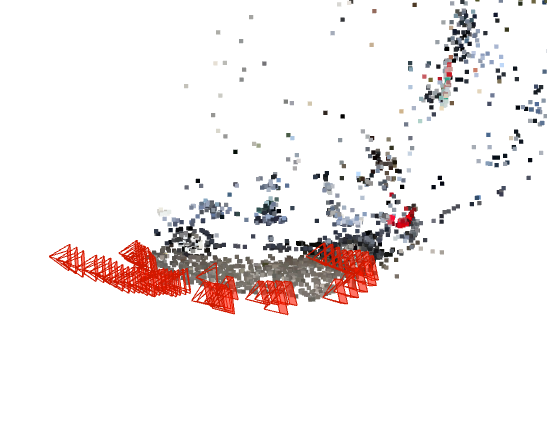} \vspace{-0.4em}
        \includegraphics[width=1\linewidth]{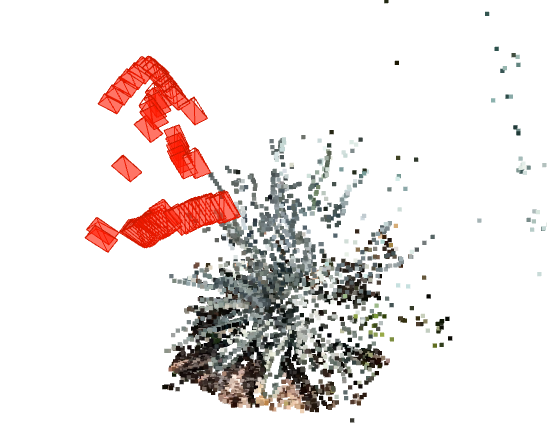} \vspace{-0.4em}
        \includegraphics[width=1\linewidth]{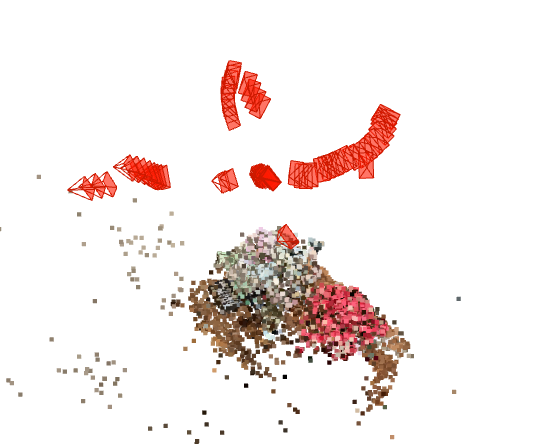} \vspace{-0.4em}
        \includegraphics[width=1\linewidth]{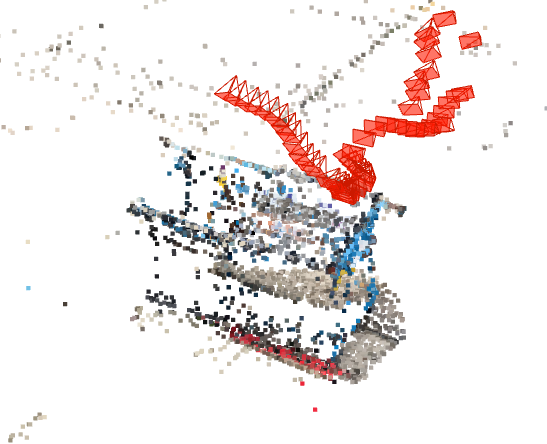} \vspace{-0.4em}
        \includegraphics[width=1\linewidth]{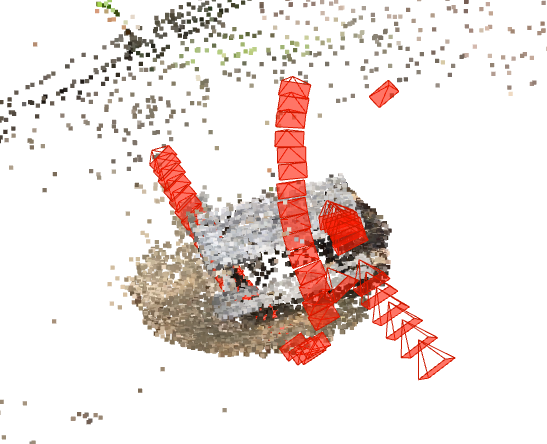} \vspace{-0.4em}
        \includegraphics[width=1\linewidth]{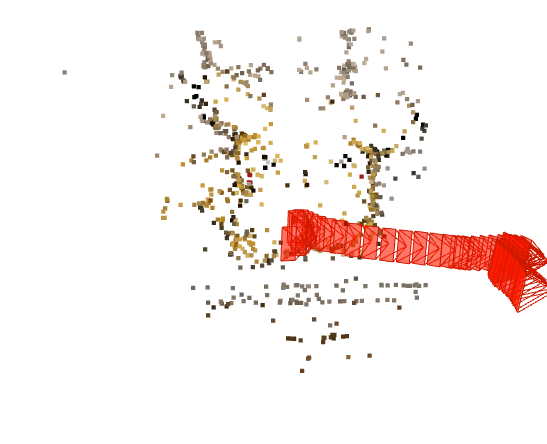} \vspace{-0.4em}
        \includegraphics[width=1\linewidth]{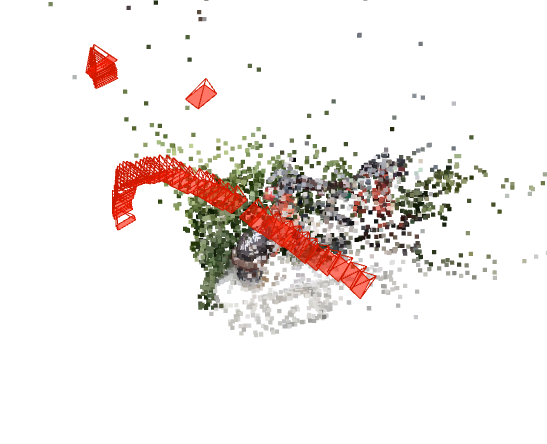} \vspace{-0.4em}
        \includegraphics[width=1\linewidth]{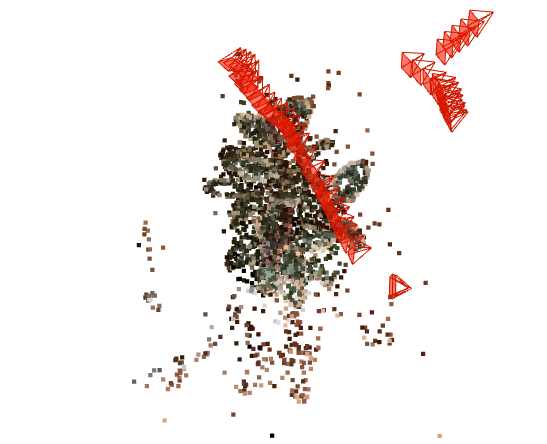} \vspace{-0.4em}
        \includegraphics[width=1\linewidth]{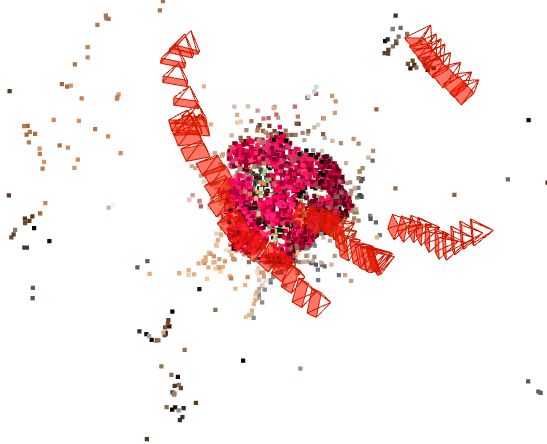} \vspace{-0.4em}
        \includegraphics[width=1\linewidth]{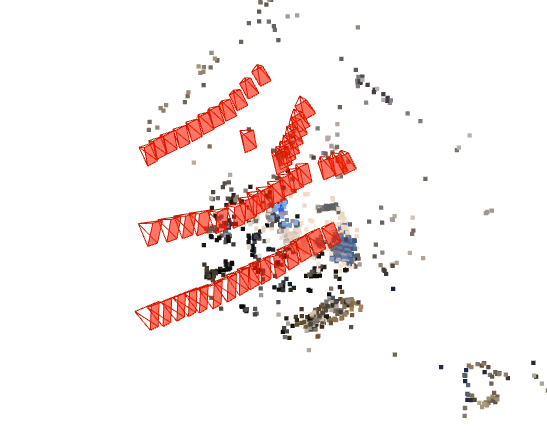}
    \end{minipage}} 
    \subfigure[CF-NeRF]{
    \begin{minipage}[b]{0.133\linewidth}
        \includegraphics[width=1\linewidth]{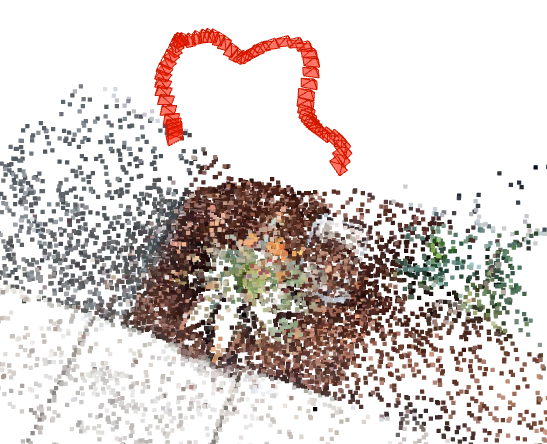} \vspace{-0.4em}
        \includegraphics[width=1\linewidth]{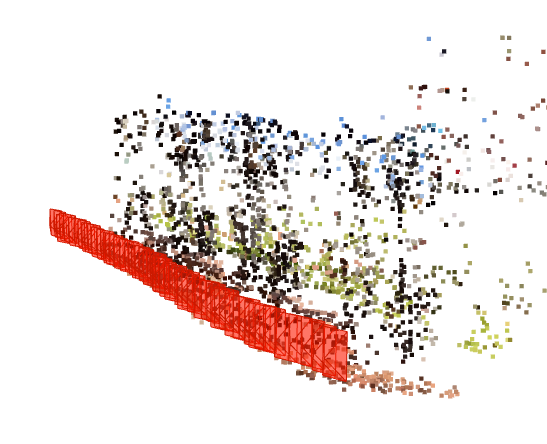} \vspace{-0.4em}
        \includegraphics[width=1\linewidth]{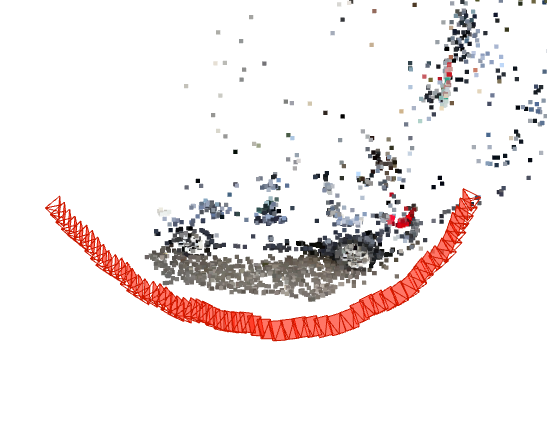} \vspace{-0.4em}
        \includegraphics[width=1\linewidth]{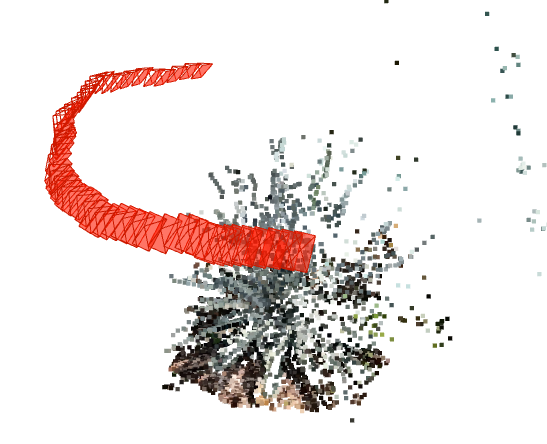} \vspace{-0.4em}
        \includegraphics[width=1\linewidth]{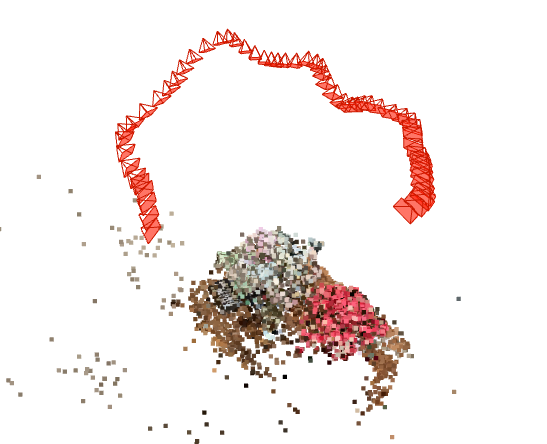} \vspace{-0.4em}
        \includegraphics[width=1\linewidth]{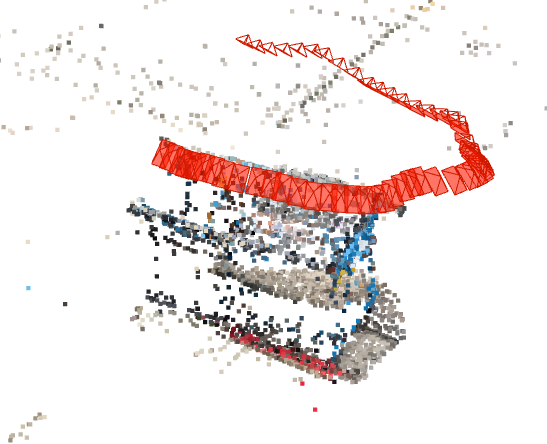} \vspace{-0.4em}
        \includegraphics[width=1\linewidth]{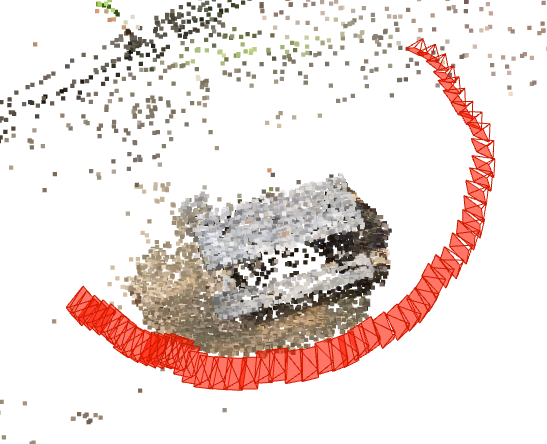} \vspace{-0.4em}
        \includegraphics[width=1\linewidth]{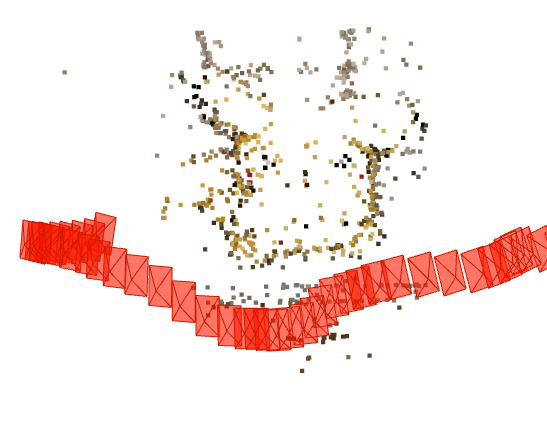} \vspace{-0.4em}
        \includegraphics[width=1\linewidth]{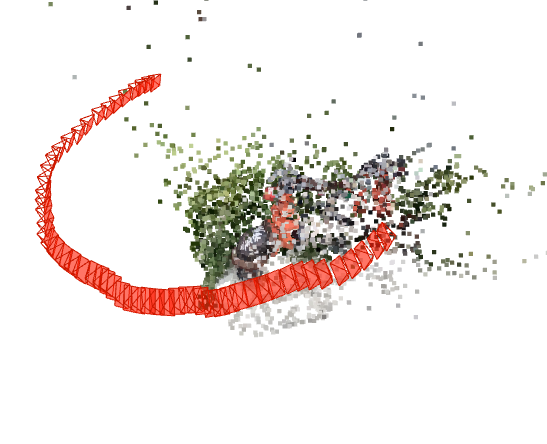} \vspace{-0.4em}
        \includegraphics[width=1\linewidth]{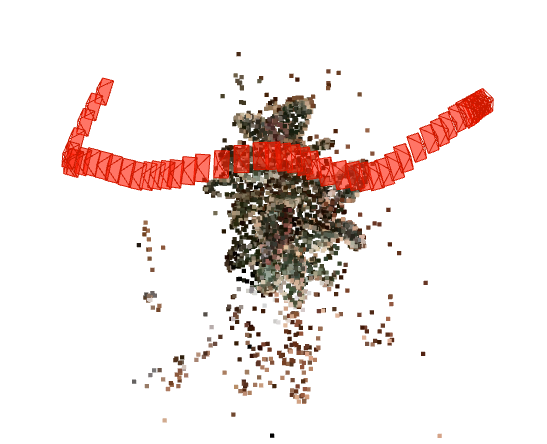} \vspace{-0.4em}
        \includegraphics[width=1\linewidth]{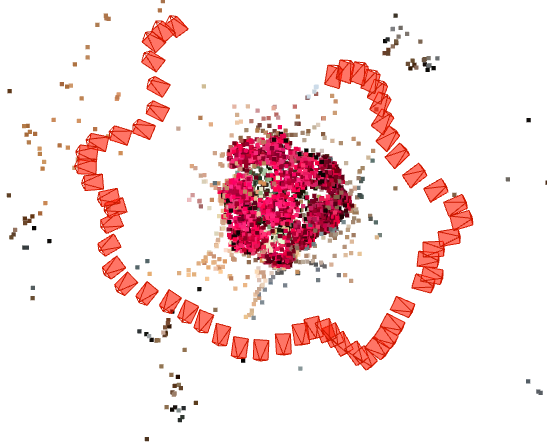} \vspace{-0.4em}
        \includegraphics[width=1\linewidth]{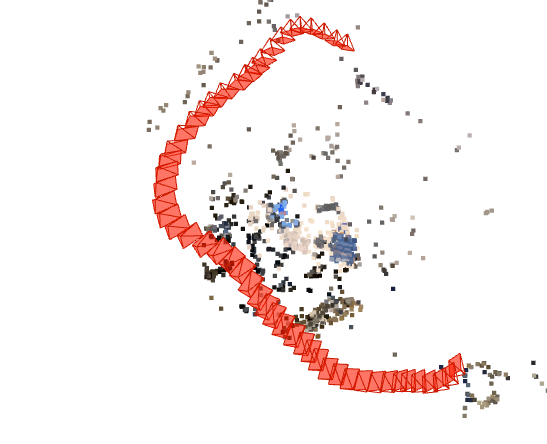}
    \end{minipage}}
    
	\vspace{-1.0em}
	
	\caption{
		\textbf{Comparison of camera parameters.} We compare camera parameters of our method CF-NeRF with NeRFmm \cite{wang2021nerf}, SiNeRF \cite{ventusff2021}, BARF \cite{lin2021barf}, GARF \cite{chng2022garf}, and L2G-NeRF \cite{chen2023local} along with sparse points from COLMAP.
	}
	\label{fig:com}
	\vspace{-2.5em}
    
\end{figure*}

\bibliography{aaai24}

\end{document}